\documentclass[nohyperref]{article}

\usepackage[preprint]{style_neurips/neurips_2023}
\usepackage[utf8]{inputenc}
\usepackage[T1]{fontenc}
\usepackage[dvipsnames]{xcolor}
\usepackage[pdftex]{ graphicx }
\usepackage{ algorithm, algorithmic }
\usepackage{ amsmath, amssymb, dsfont, mathtools, amsthm, amsfonts, nicefrac, microtype}
\usepackage{ stfloats }
\usepackage{ url }
\usepackage{ booktabs }
\usepackage{ makecell }
\usepackage{ dsfont }
\usepackage{ adjustbox }
\usepackage{ caption }
\usepackage{ subcaption }
\usepackage{ color }
\usepackage{ float }
\usepackage{natbib}
\usepackage{ hyperref }
\usepackage[capitalise, noabbrev]{ cleveref }

\theoremstyle{plain}

\theoremstyle{definition}

\theoremstyle{remark}

\usepackage[title]{appendix}
\usepackage{multirow}
\usepackage{pifont}

\DeclareMathOperator*{\argmax}{arg\,max}

\title{Intensity-free Integral-based Learning of Marked Temporal Point Processes}

\author{
  Sishun Liu \\
  STEM College \\
  RMIT University \\
  Melbourne, Victoria 3000 \\
  \texttt{sishun.liu@student.rmit.edu.au} \\
  \And 
  Ke Deng \\
  STEM College \\
  RMIT University \\
  Melbourne, Victoria 3000 \\
  \texttt{ke.deng@rmit.edu.au} \\
  \And 
  Xiuzhen Zhang \\
  STEM College \\
  RMIT University \\
  Melbourne, Victoria 3000 \\
  \texttt{xiuzhen.zhang@rmit.edu.au} \\
  \And
  Yongli Ren \\
  STEM College \\
  RMIT University \\
  Melbourne, Victoria 3000 \\
  \texttt{yongli.ren@rmit.edu.au} \\
}

\begin{document}

\maketitle

%jz1: below, the title needs more work. "harness"  = " to make use of" so not suitable. There are two "process" in the title. 

%\linespread{1.25}

%\setlength{\parindent}{0em}
%\setlength{\parskip}{0.4em}

\begin{abstract}
In the marked temporal point processes (MTPP), a core problem is to parameterize the conditional joint PDF (probability distribution function) $p^*(m,t)$ for inter-event time $t$ and mark $m$, conditioned on the history. The majority of existing studies predefine intensity functions. Their utility is challenged by specifying the intensity function's proper form, which is critical to balance expressiveness and processing efficiency. Recently, there are studies moving away from predefining the intensity function -- one models $p^*(t)$ and $p^*(m)$ separately, while the other focuses on temporal point processes (TPPs), which do not consider marks. This study aims to develop high-fidelity $p^*(m,t)$ for discrete events where the event marks are either categorical or numeric in a multi-dimensional continuous space. We propose a solution framework IFIB (\underline{I}ntensity-\underline{f}ree \underline{I}ntegral-\underline{b}ased process) that models conditional joint PDF $p^*(m,t)$ directly without intensity functions. It remarkably simplifies the process to compel the essential mathematical restrictions. We show the desired properties of IFIB and the superior experimental results of IFIB on real-world and synthetic datasets. The code is available at \url{https://github.com/StepinSilence/IFIB}.
\end{abstract}

\section{Introduction}
Events have been generated continuously in human activities or observed from natural phenomena. The events can be financial transactions, social media user activities, Web page visits by users, patient visits to clinics, earthquake occurrences in seismology, neural spike trains in neuroscience, the extreme temperature in the weather forecast, and the observations of rare birds in ecology. Temporal point processes (TPP) are generative models of variable-length point sequences which represent the arrival times of events. TPPs are built upon rich theoretical foundations, with early work dating back to many decades ago, where they were used to model the arrival of insurance claims and telephone traffic \cite{shchur_intensity-free_2020}, till now widely applied in social network analysis, neural logic inference, and biological activity modeling.

The marked TPP (MTPP) concerns the scenarios where each event comes with an arrival time as well as a mark. The mark can be categorical such as the magnitude of earthquakes, mild/moderate/critical symptoms of patients visiting an emergency, sell/buy in financial transactions, or numeric such as temperature in the weather forecast, the longitude and latitude of observations in ecology. As often encountered in practice, the marked TPP has attracted much attention from the research community \cite{du_recurrent_2016, mei_neural_2017, guo_initiator_2018, enguehard_neural_2020, zuo_transformer_2020, zhang_self-attentive_2020, mei_noise-contrastive_2020, shchur_intensity-free_2020, Marin2019, chen_neural_2020}. Most studies assume the events in sequence are correlated so that the marked TPPs are conditioned on history, i.e., the events that occurred so far.  

A core problem of marked TPP is to parameterize the conditional joint PDF $p^*(m, t)$\footnote{The asterisk reminds us the probability is conditioned on history.} for inter-event time $t$ and mark $m$, conditioned on the history $\mathcal{H}_{t_l}$ \cite{shchur_intensity-free_2020}. Many applications regarding the time and mark of the next event depend on $p^*(m,t)$. The widely studied one is to predict when the next event will occur and, given the time, which mark the next event is \cite{du_recurrent_2016, mei_neural_2017, guo_initiator_2018, enguehard_neural_2020, zuo_transformer_2020, zhang_self-attentive_2020, mei_noise-contrastive_2020, shchur_intensity-free_2020}. Moreover, for each mark, if its probability to be the next event is non-zero, it is interesting to predict when the next event will occur conditioned on the fact that the next event is the mark. Besides that, one can report the evolution of probabilities for different marks to be the next event over time \cite{Marin2019}.

The majority of existing studies model $p^*(m, t)$ by defining the intensity function \cite{enguehard_neural_2020, Daryl2003, mei_neural_2017, zuo_transformer_2020}. They suffer from the expressiveness issue if the intensity function is simple or encounter the computationally-expensive intensity integral problem if complex. One recent study moves away from specifying the form of the intensity function by exploring a neural network \cite{omi_fully_2019}. The solution known as FullyNN is designed for TPP rather than marked TPP, which does not consider event marks. In another recent study, an intensity-free solution has been proposed \cite{shchur_intensity-free_2020} where $p^*(m)$ and $p^*(t)$ are modeled separately, still, it does not directly address the challenge of modeling $p^*(m,t)$. The intensity function of TPP where marks are locations in a continuous spatial space has been studied recently \cite{chen_neural_2020}.       

%To model $p^*(m,t)$ for the marked TPP, a natural solution is to extend FullyNN to include information about event marks. This solution, namely \underline{F}ully \underline{E}vent \underline{N}eural \underline{N}etwork (FENN). We show that FENN is a feasible solution, but it inherits the intrinsic shortcomings of FullyNN -- estimates of malformed probability distributions. 

This study aims to develop high-fidelity $p^*(m,t)$ for discrete events where the event marks are either categorical or numeric. If numeric, the event mark is represented as a vector in a multi-dimensional continuous space. We propose a solution framework IFIB (\underline{I}ntensity-\underline{f}ree \underline{I}ntegral-\underline{b}ased process) that models conditional joint PDF $p^*(m,t)$ directly without intensity functions. It remarkably simplifies the process to compel the essential mathematical restrictions. IFIB has two variants, IFIB-C and IFIB-N, where IFIB-C is IFIB for \underline{c}ategorical marks, and IFIB-N is IFIB for \underline{n}umeric marks. To the best of our knowledge, IFIB is the first model of its kind. The superiority of IFIB has been verified by experiments on real-world and synthetic datasets in different applications. The source code and data used in this study will be available upon acceptance.

%To enable high-fidelity models of discrete events that are either categorical or numeric localized in continuous time and space. imporove We further propose 
%\underline{I}ntensity-\underline{f}ree \underline{I}ntegral-\underline{b}ased (IFIB) process to directly model the integral of $p^*(m,t)$. Specifically, IFIB explores the relationship between $p^*(m, t)$ and $p^*(m)$, the marginal distribution of $m$, rather than involving intensity function as in FENN. Our study shows IFIB has the desired properties that overcome the shortcomings of FENN. The superiority of IFIB has been verified by experiments on real-world and synthetic datasets in different applications.  

\section{Preliminaries}
The marked TPP is a random process whose embodiment is a sequence of discrete events, $\mathcal{S} = \{(\mathtt{m}_i,t_i)\}_{i=1}^l$, where $i\in \mathbb{Z}^+$ is the sequence order, $t_i\in \mathbb{R}^+$ is the time when the $i$th event occurs, $\mathtt{m}_i$ is the mark of the $i$th event, and $t_i<t_j$ if $i<j$. This paper considers the simple marked TPP, which allows at most one event at every time. The time of the most recent event is $t_l$, and the current time is $t>t_l$. The time interval between two adjacent events is inter-event time. We assume that the occurrence of an event with a particular mark at a particular time may be triggered by what happened in the past. Let \(\mathcal{H}_{t_l}\) be the history up to (including) the most recent event, and \(\mathcal{H}_{t-}\) be the history up to (excluding) the current time \cite{rasmussen_lecture_2018}. In different application scenarios, the mark $\mathtt{m}$ can be either categorical or numeric. If categorical, the mark $\mathtt{m}$ is denoted as $m$ which is in a finite set of labels $\mathrm{M}=\{k_1,k_2,\cdots,k_{|\mathrm{M}|}\}$. If numeric, the mark $\mathtt{m}$ is denoted as $\mathbf{m}$ which is a vector $(d_1,\cdots,d_n)$ in $\mathbf{M}$, a $n$-dimensional continuous space where the value range is $[a_i,b_i]$ for dimension $i$. 

\textbf{Categorical Mark} when $\mathtt{m}$ is the categorical mark, the conditional intensity function of the marked TPP can be defined:% in \cref{eqn:1}:%\cite{daley_early_2003}:
\begin{equation}
%\begin{aligned}
    \lambda^*(m = k_i,t) = \lambda(m = k_i,t|\mathcal{H}_{t-}) = \lim_{\Delta t \rightarrow 0}\frac{P(m = k_i, t \in [t, t+\Delta t)|\mathcal{H}_{t-})}{\Delta t}.
    \label{eqn:1}
%\end{aligned}
\end{equation} 
With $\lambda^*(m, t)$, the conditional joint PDF of the next event can be defined:
\begin{equation}
%\begin{aligned}
p^*(m, t) = p(m,t|\mathcal{H}_{t_l}) = \lambda^*(m, t) F^*(t) = \lambda^*(m, t)\exp({-\int_{t_{l}}^t{\sum_{n\in \mathrm{M}}\lambda^*(n, t)d\tau}}).
\label{eqn:mtpp}
%\end{aligned}
\end{equation}
where $F^*(t)$ is the conditional probability that no event has ever happened up to time $t$ since $t_l$.

\textbf{Numeric Mark} When $\mathtt{m}$ is the numeric mark, the conditional intensity function of the marked TPP can be defined:
\begin{equation}
%\begin{aligned}
    \lambda^*(\mathbf{m},t) = \lambda(\mathbf{m},t|\mathcal{H}_{t-}) =\lim_{\Delta t \rightarrow 0, |\mathcal{C}(\mathbf{m})| \rightarrow 0}\frac{P(\mathbf{m} \in \mathcal{C}(\mathbf{m}), t \in [t, t+\Delta t)|\mathcal{H}_{t-})}{\Delta t |\mathcal{C}(\mathbf{m})|}.
    \label{eqn:1_c}
%\end{aligned}
\end{equation}
where \(\mathcal{C}(\mathbf{m})\) refers to a small hypercube in $\mathbf{M}$ that centers at \(\mathbf{m}\) and \(|\mathcal{C}(\mathbf{m})|\) is the volume of \(\mathcal{C}(\mathbf{m})\). With $\lambda^*(\mathbf{m}, t)$, the conditional joint PDF of the next event can be defined:
\begin{equation}
p^*(\mathbf{m}, t) = p(\mathbf{m},t|\mathcal{H}_{t_l}) = \lambda^*(\mathbf{m}, t) F^*(t) 
= \lambda^*(\mathbf{m}, t)\exp({-\int_{t_{l}}^t{\int_{\mathbf{m}\in \mathbf{M}}\lambda^*(\mathbf{m}, t)d\tau}d\mathbf{m}}).
\label{eqn:mtpp_c}
%\end{aligned}
\end{equation}
The detailed elaboration about how we obtain \cref{eqn:mtpp} from \cref{eqn:1} and obtain \cref{eqn:1_c} from \cref{eqn:mtpp_c} is in \cref{appendix:mtpp_pdf}. In this study, our task is to model the conditional joint PDF $p^*(\mathtt{m}, t)$ where $\mathtt{m}$ is either categorical or numeric. 

The simplest form of TPP is the homogeneous Poisson process whose intensity merely contains a positive number \(\lambda^*(t) = c\). It is widely used in the thinning algorithm \cite{ogata_lewis_1981} for creating synthetic datasets based on almost any predefined intensity functions. Another example is the Hawkes process \cite{hawkes_spectra_1971}, belonging to the self-exciting point process family. Its conditional intensity function is in the form of $\lambda^*(t) = \mu + \sum_{i:t_i<t}{\kappa(t, t_i)}$,  $\kappa(t, t_i) > 0$, showing that every event excites the intensity function before it falls. Because it meets the real-world intuition that people's interest always drastically drops as time passes, the Hawkes process is a widely used prior distribution in various TPP models \cite{cao_deephawkes_2017, mei_neural_2017, salehi_learning_2019}.

\section{Related Work}
Most studies specify a separate intensity function for each categorical mark $k$ (i.e., the probability that the event of a particular mark will occur at any specific future time) based on which density function $p^*(m, t)$ can be formulated \cite{enguehard_neural_2020, Daryl2003, mei_neural_2017, zuo_transformer_2020, du_recurrent_2016}. All these solutions assume a specific functional form and require the intensity integral to derive the density function. As pointed out in \cite{shchur_intensity-free_2020}, this is usually considered their intrinsic shortcomings due to the trade-off between efficiency and effectiveness. For a "simple" intensity function like in \cite{du_recurrent_2016}, the intensity integral has a closed form, which makes the log-likelihood easy to compute. However, such models usually have limited expressiveness. A more sophisticated intensity function like in \cite{mei_neural_2017} can better capture the dynamics of the system, but computing log-likelihood will require approximating the intensity integral using a numerical method such as Monte Carlo. 
 
Recent studies~\cite{shchur_intensity-free_2020, omi_fully_2019,chen_neural_2020} move away from predefining intensity functions. In \cite{shchur_intensity-free_2020}, an intensity-free solution has been proposed to infer $p^*(t)$ from a simple distribution such as standard normal distribution or mixture Gaussian via a stack of differentiable invertible transformations. In the scenarios of multiple marks, the intensity-free solution factorizes $p^*(m, t)$ into a product of two independent distributions $p^*(t)$ and $p^*(m)$. Even though it is conceptually possible to provide $p^*(m,t)$, it remains a challenge to ensure that the $p^*(m,t)$ integral across all marks is 1, an essential property of probability density distribution. In \cite{omi_fully_2019}, a method known as FullyNN has been proposed to model the intensity integral using a neural network from which the intensity can be derived by differentiation, an operation computationally much easier compared with integral. FullyNN was proposed for TPP rather than marked TPP, which does not consider event marks. Also, FullyNN cannot guarantee essential mathematical restrictions \cite{shchur_intensity-free_2020}. In \cite{chen_neural_2020}, the spatio-temporal point processes are investigated by leveraging Neural ODEs as the computational method for modeling discrete events where the marks are locations in a continuous spatial space. It concerns intensity function modeling only.    

\section{Method}\label{sec:method}
We propose a solution framework IFIB (\underline{I}ntensity-\underline{f}ree \underline{I}ntegral-\underline{b}ased process) that models the relationship between $p^*(\mathtt{m},t)$ and its integral. The most relevant method to IFIB is FullyNN \cite{omi_fully_2019} that models the relationship between $\lambda^*(\mathtt{m},t)$ and its integral. IFIB constructs $p^*(\mathtt{m},t)$ directly instead of using $\lambda^*(\mathtt{m},t)$. As a result, IFIB remarkably simplifies the process to compel the essential mathematical restrictions. IFIB has two variants, IFIB-C and IFIB-N, where IFIB-C is IFIB for \underline{c}ategorical marks, and IFIB-N is IFIB for \underline{n}umeric marks.  %FullyNN considers all events to have the same mark \footnote{Appendix B.1 explains why FullyNN is incapable of MTPP modeling and an intuitive solution based on FullyNN is introduced in \cref{apx:ETE}.}. 

\subsection{IFIB-C}\label{sec:IFIBP}
\begin{figure*}[ht]
    \centering
    \includegraphics[width=0.95\linewidth]{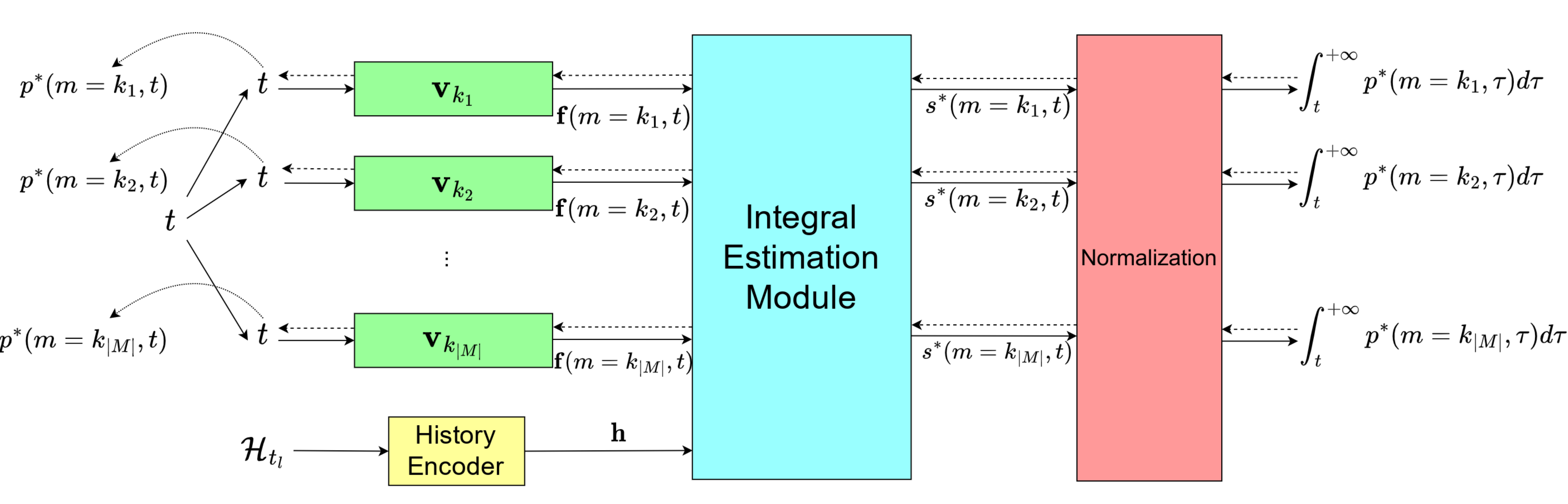}
    \caption{Architecture of IFIB-C. The solid arrows refer to forward propagation, the dashed arrows refer to backpropagation, and the curved dotted arrows refer to retrieving the gradient. The history encoder is an LSTM.}
    \label{fig:IFIB}
\end{figure*}

\cref{fig:IFIB} sketches the architecture of IFIB-C. Given a categorical mark $m\in \mathrm{M}$, IFIB-C explores the marginal probability distribution of $m$, $p^*(m)$, which is the integral of $p^*(m,t)$ over time from the last event as shown in \cref{eqn:mt_m_dist}. 
\begin{equation}\label{eqn:mt_m_dist}
p^*(m) = \int_{t_l}^{+\infty}{p^*(m, \tau)d\tau} 
\end{equation}

For each mark $m$, we assign a vector $\mathbf{v}_m$ to prepare $\mathbf{f}(m, t)=\mathbf{v}_{m}(t - t_l)$ as input of the integral estimation module (IEM). IEM contains multiple fully-connected layers with non-negative weights and monotonic-increasing activation functions. It ends with a monotonically-decreasing sigmoid function $\sigma^{\prime}(x) = 1/(1 + e^x)$ for each mark. 

The outputs of IEM are scores $s^*(m = k_1, t), s^*(m = k_2, t), \cdots, s^*(m = k_{|\mathrm{M}|}, t)$. The value of $\sum_{m\in \mathrm{M}}s^*(m,t)$ is not guaranteed to be 1. In order to produce the qualified probability distribution, they need to be normalized. This is achieved by Normalization module in \cref{fig:IFIB} that divides $s^*(m, t)$ by the partition function $Z(\mathcal{H}_{t_l}) = \sum_{m \in \mathrm{M}}{s^*(m, t_l)}$ for each $m\in M$. Finally, IFIB outputs $\Gamma^*(m,t)$ for each mark $m$ at the given time $t$:

\begin{align}
\label{eqn:IFIB_1}
    \Gamma^*(m, t) &=\int_{t}^{+\infty}{p^*(m, \tau)d\tau} = \frac{s^*(m, t)}{Z(\mathcal{H}_{t_l})} \\
% \end{align}
% \begin{align}
\label{eqn:IFIB_2}
    p^*(m, t) %= - \frac{1}{Z(\mathcal{H}_{t_l})}\frac{\partial p^*(m)}{\partial t} \\
    &= -\frac{1}{Z(\mathcal{H}_{t_l})}\frac{\partial \Gamma^*(m, t)}{\partial s^*(m, t)}\frac{\partial s^*(m, t)}{\partial \mathbf{f}(m, t)}\frac{\partial \mathbf{f}(m, t)}{\partial t}
\end{align}
Note that $p^*(m)$ in \cref{eqn:mt_m_dist} and $\Gamma^*(m, t)$ are distinct integrals. The former starts from \(t_l\), the time of the last event in history, while the latter starts from time \(t\), any time after $t_l$. When $t=t_l$, $p^*(m)$ is equivalent to $\Gamma^*(m, t)$ and $\sum_{m\in \mathrm{M}}p^*(m)=\sum_{m\in \mathrm{M}}\Gamma^*(m, t)=1$. The loss function of IFIB is \cref{eqn:mtpp_loss} that is the sum of the negative log-likelihood of $p^*(m, t)$ at every event $(m_i,t_i)\in \mathcal{S}$. 
\begin{equation}\label{eqn:mtpp_loss}
%\begin{aligned}
    % L = -\sum_{(m_i,t_i)\in \mathcal{S}} \log p^*(m_i,t_i)+\sum_{m'\in M}{\mathds{1}_{m_i}\log(p^*(m'))}. \\
    L = -\sum_{(m_i,t_i)\in \mathcal{S}} \log p^*(m_i,t_i). 
    % \\
    % L &= \sum_{(m, t)\in \mathcal{S}} -\log p^*(m,t) + \operatorname{cross\_entropy}(p, m) \\
    % \operatorname{cross\_entropy}(p^*(m), m) &= - \sum_{m \in M}{p_t(m|\mathcal{H})\log(p(m|\mathcal{H}))} \\ 
%       &= -\log (-\frac{\partial \operatorname{IFIB}(m, t, \mathcal{H}_{t_l})}{\partial t}) - \log(\operatorname{IFIB}(m, t_l, \mathcal{H}_{t_l}))
% \end{aligned}
\end{equation}
where $p^*(m_i,t_i)$ is the predicted probability after $(i-1)$th event.  

\subsection{IFIB-N}\label{sec:cifib}
This section introduces IFIB-N, an IFIB variant for marked TPP where each mark is a vector in an $n$-dimensional continuous space, denoted as $\mathbf{m}=(d_1,d_2,\cdots,d_n)$. IFIB-N outputs $\Gamma^*(\mathbf{m}, t)$, the integral of $p^*(\mathbf{m},t)$ over time from $t$ and over n-dimensional continuous space from $\mathbf{m}$:   
\begin{equation}
\begin{aligned}
\label{eqn:cifib_1}
    \Gamma^*(\mathbf{m}, t) %&= \int_{t}^{+\infty}{\int_{\mathbf{D_m}}{p^*(\mathbf{m}, \tau)d\tau}d\mathbf{D_m}} \\ 
    &= \int_{\tau=t}^{+\infty}{\int_{r_1=d_1}^{b_1}\int_{r_2=d_2}^{b_2}\cdots \int_{r_n=d_n}^{b_n}{p^*(r_1,r_2,\cdots,r_n,\tau)d\tau}dr_1dr_2\cdots dr_n}
\end{aligned}
\end{equation}
where \(p^*(\mathbf{m}, t)\) is defined as follows:
\begin{equation}
%\begin{aligned}
\label{eqn:cifib_2}
    p^*(\mathbf{m}, t) = (-1)^{n + 1} \frac{\partial}{\partial t}(\frac{\partial}{\partial d_1}(\frac{\partial}{\partial d_2}(\cdots(\frac{\partial \Gamma^*(\mathbf{m}, t)}{\partial d_n})))) \\
    %&= (-1)^{n + 1} \frac{\partial}{\partial t}(\frac{\partial^{n} \Gamma^*((m_1, m_2, \cdots, m_n), t)}{\partial m_1\partial m_2\cdots \partial m_n})
%\end{aligned}
\end{equation}
However, we cannot devise an end-to-end model for estimating \(\Gamma^*(\mathbf{m}, t)\) as \cref{eqn:cifib_2} because such model's (\(n+1\))-rank derivative must be non-negative. This restriction eliminates most candidate functions while the remaining ones that could fulfill the restriction, like the exponential functions, are prone to enlarge the output and finally lead to the output explosion. Therefore, we propose a trick to split \(\Gamma^*(\mathbf{m}, t)\) into the product of integrals of \(n + 1\) conditional probability distributions, shown in \cref{eqn:ifib_n_target}, each of which can be estimated separately. This trick dispenses the \(n + 1\) differentiation operations into the \(n + 1\) inputs, i.e., one differentiation operation for each input. Thus, we only need to affirm that the selected model could always estimate a normalized probability distribution.%, which IFIB has already satisfied.
\begin{equation}
\begin{aligned}
    \Gamma^*(\mathbf{m}, t) %&= \int_{t}^{+\infty}{\int_{\mathbf{m}}^{(b_1, b_2, \cdots, b_n)}{p^*(\mathbf{r}, \tau)d\tau}d\mathbf{r}} \\
    &= \int_{\tau=t}^{+\infty}{p^*(\tau)}d\tau{\int_{d_1}^{b_1}{p^*(r_1|\tau)}dr_1{\cdots \int_{d_n}^{b_n}{p^*(r_n|\tau, r_1, r_2, \cdots, r_{n-1})}}}dr_{n}
    \label{eqn:ifib_n_target}
\end{aligned}
\end{equation}

\begin{figure}
    \centering
    \includegraphics[width=\textwidth]{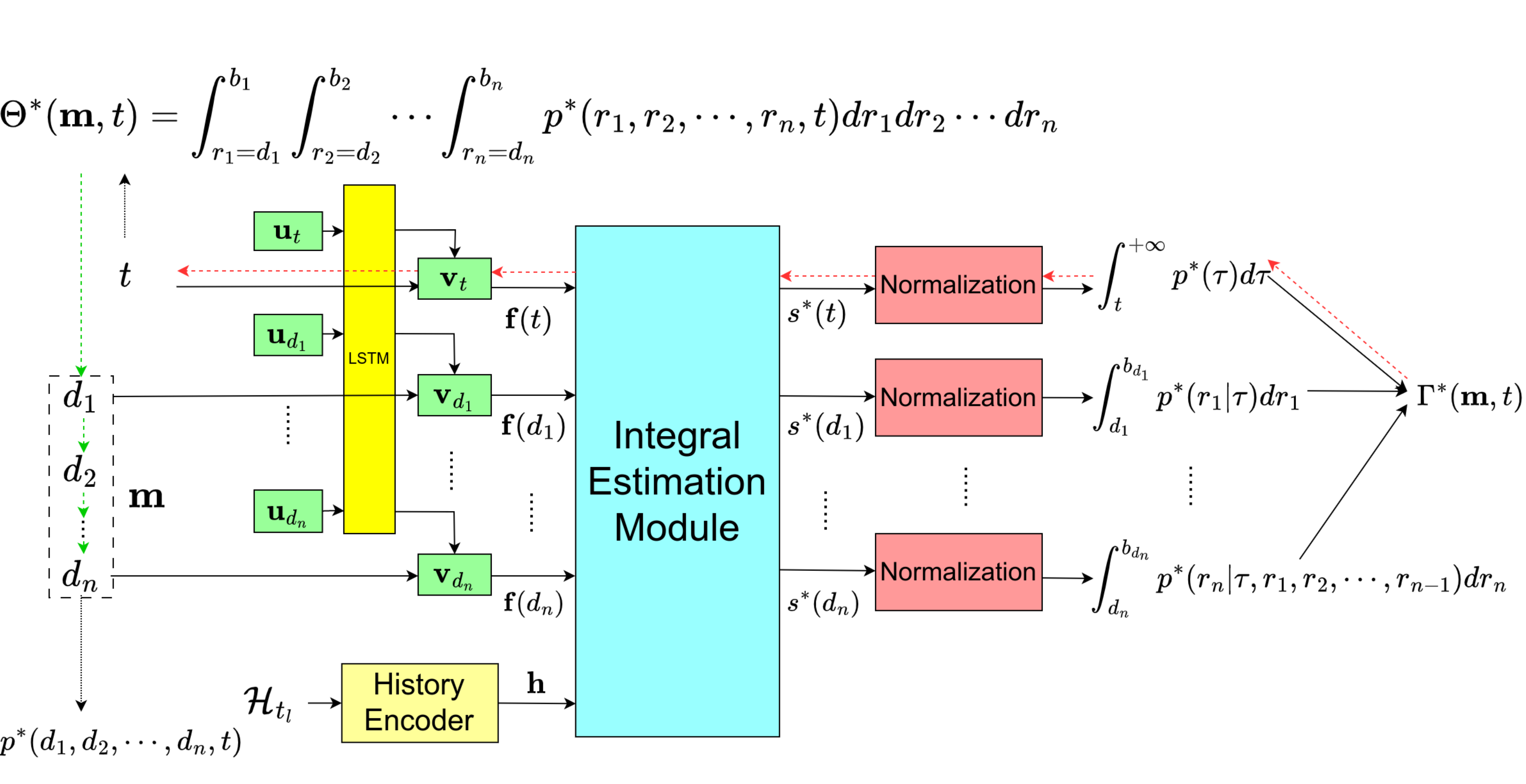}
    \caption{Architecture of IFIB-N. The solid, dashed, and dotted arrows are the same as in \cref{fig:IFIB}. The history encoder is an LSTM.}% By splitting the target integral into the product of integrals of conditional probability distributions, CIFIB enables learning the multiple integral of a joint probability distribution \(p(\mathbf{m}, t)\) in a specific high-dimensional space.}
    \label{fig:cifib}
\end{figure}

% We might leave the deep reason about CIFIB's kind-of redundant design in the Appendix?
% No.
\cref{fig:cifib} depicts the structure of IFIB-N. For time \(t\) and each dimension $i$ of mark \(\mathbf{m}\), we allocate exclusive embedding vector \(\mathbf{u}_{t}\) and \(\mathbf{u}_{d_i}\). An LSTM module and a non-negative activation function are applied to produce non-negative \(\mathbf{v}_{t}\) and \(\mathbf{v}_{d_i}\) from \(\mathbf{u}_{t}\) and \(\mathbf{u}_{d_i}\) to disseminate conditional information to each probability distribution in \cref{eqn:ifib_n_target}. The dot product of \(t\) with \(\mathbf{v}_{t}\) is $\mathbf{f}(t)$ which represents $p^*(\tau)$. The dot product of \(d_i\) with \(\mathbf{v}_{d_i}\) is $\mathbf{f}(d_i)$ where $\mathbf{f}(d_1)$, $\cdots$, $\mathbf{f}(d_n)$ represent $p^*(r_1|\tau)$, $\cdots$, $p^*(r_n|\tau,r_1,\cdots,r_{n-1})$, respectively.
% connect all embedding vectors to enable information sharing among all embedding vectors.
% IFIB-N encodes information about every conditional probability in \cref{eqn:split} into these embedding vectors. Hence, 
% we apply an LSTM module to connect all embedding vectors to enable information sharing among all embedding vectors. 
The IEM stays the same as in IFIB-C. For each of IEM outputs \(s^*(t)\) and \(s^*(d_i)\), it has a Normalization module as in IFIB-C to ensure the output integral is normalized. Then, we multiply the integral of all conditional probabilities to form \(\Gamma^*(\mathbf{m}, t)\). 

IFIB-N owns a unique backpropagation process to obtain \(p^*(\mathbf{m}, t)\). We elaborate this process in \cref{alg:cifib_get_p}. Briefly, it comprises two steps. First, we differentiate the final output by time \(t\)(represented by red dashed arrows in \cref{fig:cifib}) for \(\Theta^*(\mathbf{m}, t)\). Then, we obtain \(p^*(\mathbf{m}, t)\) by differentiating \(\Theta^*(\mathbf{m}, t)\) with every dimension of \(\mathbf{m}\) (represented by grass green dashed arrows in \cref{fig:cifib}). 

\begin{algorithm}[tb]
\caption{Obtain \(p^*(\mathbf{m}, t)\) from \(\Gamma^*(\mathbf{m}, t)\) }
\label{alg:cifib_get_p}
\begin{algorithmic}
\STATE {\bfseries Input:} Multi-dimensional Integral \(\Gamma^*(\mathbf{m}, t)\), input time \(t\), input mark \(\mathbf{m} = (d_1, d_2, \cdots, d_n)\);
\STATE \(\Theta^*(\mathbf{m}, t) = \partial \Gamma^*(\mathbf{m}, t)/\partial t \);
\STATE grad = \(\Theta^*(\mathbf{m}, t)\);
\FOR {\(d_i\) {\bfseries in} \(\mathbf{m} = (d_1, d_2, \cdots, d_n)\)}
\STATE grad = \( \partial \text{grad}/\partial d_i \);
\ENDFOR
\STATE \(p^*(\mathbf{m}, t) = \text{grad}\);
\RETURN \(p^*(\mathbf{m}, t)\)
\end{algorithmic}
\end{algorithm}

% \subsection{Desired Properties of IFIB}
\subsection{Why integral of distribution from \(t\) to infinity?}
% FullyNN estimates the integral of the intensity function \(\lambda^*(t)\) from \(t_l\) to \(t\). However,
Both IFIB-C and IFIB-N involve the integral of \(p^*(\mathtt{m}, t)\) from \(t\) to positive infinity. One might raise questions about this design: what is the advantage of estimating the integral of \(p^*(\mathtt{m}, t)\) from \(t\) to positive infinity? Why does not IFIB estimate the integral of \(p^*(\mathtt{m}, t)\) from \(t_l\) to \(t\)?
Suppose we choose to estimate the integral of \(p^*(\mathtt{m}, t)\) from \(t_l\) to \(t\). Our model \(\mathcal{M}\) must satisfy two restrictions: (1) output 0 when the input is \(t_l\) as \(\mathcal{M}(t_l) = \int_{t_l}^{t_l}{p^*(\mathtt{m}, \tau)d\tau} = 0\), and (2) output no bigger than 1 as \(t\) increases because \(\lim_{t \rightarrow +\infty}{\mathcal{M}(t)} = \int_{t_l}^{+\infty}{p^*(\mathtt{m}, \tau)d\tau} \in [0, 1]\) \cite{shchur_intensity-free_2020}. While the latter is relatively easy, achieving the former is difficult since the output must be 0 when the input is \(t_l\) regardless of the model parameter. 

If we estimate the integral of \(p^*(\mathtt{m}, t)\) from \(t\) to positive infinity, both restrictions are transformed into different restrictions. That is, our model \(\mathcal{M}\) must (1) start from a positive number in \([0, 1]\) because \(\mathcal{M}(t_l) = \int_{t_l}^{+\infty}{p^*(\mathtt{m}, \tau)d\tau} \in [0, 1]\), and (2) converge to 0 as \(t\) goes larger because \(\lim_{t\rightarrow +\infty}{\mathcal{M}(t)} = \lim_{t\rightarrow +\infty}\int_{t}^{+\infty}{p^*(\mathtt{m}, \tau)d\tau} = 0\). Such two restrictions are much easier for a model to comply with. %Thus, we decide that IFIB-C and IFIB-N should estimate \(\Gamma^*(\mathtt{m}, t)\), which helps IFIB because the first integral-based MTPP model that always estimates normalized probability functions.

\subsection{Applications}
\label{sec:application}
Given IFIB-C and IFIB-N, different applications regarding the mark and time of the next event can be performed. %\footnote{For simplicity, we do not replace \(\int_{t}^{+\infty}{p^*(m, \tau)d\tau}\) in the equations below with \(\frac{\operatorname{IFIB}(m, t, \mathcal{H}_{t_l})}{\sum_{n \in M}{\operatorname{IFIB}(n, t_l, \mathcal{H}_{t_l})}}\) }. 
The widely studied application is to predict when the next event will occur and, once the time is determined, to predict which mark the next event is. We name it \textbf{time-event prediction problem}. Moreover, for each mark, if its probability of being the next event is non-zero, it is interesting to predict when the next event will occur conditioned on the fact that the next event is the mark. We name it \textbf{event-time prediction problem}. We explain how IFIB-C and IFIB-N solve these tasks in \cref{app:solve_two_tasks}.

\section{Experiments on IFIB-C}
In this section, we evaluate IFIB-C and baselines on four real-world datasets including Bookorder (BO) \cite{du_recurrent_2016}, Retweet \cite{zhao_seismic_2015}, StackOverflow (SO) \cite{Leskovec2014SNAPD}, and MOOC \cite{shchur_intensity-free_2020}) and five synthetic datasets including Hawkes\_1, Hawkes\_2, Poisson, Self-correct, and Stationary Renewal \cite{omi_fully_2019}. Detailed information about these datasets is available in \cref{app:datasets}.

% \subsubsection{Synthetic Datasets}
%  We generated five synthetic datasets(hawkes\_1, hawkes\_2, poisson, self-correct, and stationary-renewal) following \cite{omi_fully_2019}. Each synthetic dataset has 160,000 sequences for training, 20,000 for evaluation, and 20,000 for testing, and each sequence has 64 events with five marks generated by a uniform distribution. Detailed configurations about these synthetic datasets are described in \cref{app:synthetic_generation}.

\subsection{Evaluation Metrics}
\label{sec:em}
For real-world datasets, we utilize MAE (\underline{M}ean \underline{A}bsolute \underline{E}rror) and macro-F1 for the time-event prediction problem and MAE-E (\underline{M}ean \underline{A}bsolute \underline{E}rror by \underline{E}vent) and macro-F1 for the event-time prediction problem. To measure more reliably, we sort the prediction errors and report Q1, median, Q3 (i.e., $25$th, $50$th, $75$th percentile), denoted as MAE$@25\%$, MAE$@50\%$, MAE$@75\%$, respectively. We do the same for MAE-E. For the synthetic datasets, Spearman's coefficient, \(L^1\) distance, and the relative NLL loss are selected to gauge the difference between the learned \(\hat{p}^*(m, t)\) and the real \(p^*(m, t)\) as in \cite{omi_fully_2019, shchur_intensity-free_2020}. Details of evaluation metrics are available in \cref{app:metric}.

%\textbf{Evaluation Metric}
%We use classic MAE for distribution evaluation, Spearman coefficient \(\rho\) and \(L^1\) distance for measuring the similarity between learned intensity functions and ground truth for synthetic datasets. As for marked real-world datasets, we provide MAE per event(pe-MAE) measuring model prediction performance and macro-F1 for marker prediction performance. Detailed descriptions of the used metrics are available in \cref{app:metric}.\par

%\textbf{Hyperparameters} Detailed information about hyperparameters is available in \cref{app:hyperparameters}. We have carefully tuned all hyperparameters so that our models do not overfit on training sets. \par

%\textbf{Model Settings} Shchur et al.notice that FullyNN suffers from several theoretical issues\cite{shchur_intensity-free_2020} including faulty integral estimation. Our experiments results also prove that such faulty estimation could significantly damage the event prediction performance with tiny training sets. To tackle this, we compare Enguehard et al.'s offset trick\cite{enguehard_neural_2020}(we add suffix "+ZS" when using this trick) with adding a contrastive loss \(L_{n}\) to the MTPP loss in \cref{eqn:mtpp_loss}(we add suffix "+CL" when using this trick), and find both tricks could help two proposed MTPP models recover to a reasonable event prediction accuracy. Detailed analysis about listed two tricks are available in \cref{app:tricks}.

\subsection{Baseline Models}
%We compare IFIB, FENN, and several baselines.
In this paper, we select six classic neural temporal point process models, Recurrent Marked Temporal Point Process(RMTPP) \cite{du_recurrent_2016}, Fully Neural Network(FullyNN)\cite{omi_fully_2019}, Fully Event Neural Network (FENN), Transformer Hawkes Process (THP)\cite{zuo_transformer_2020}, LogNormMix\cite{shchur_intensity-free_2020}, and Self-Attentive Hawkes Process (SAHP)\cite{zhang_self-attentive_2020} as baselines. Detailed information about these approaches is available in \cref{app:baseline}.

\subsection{Experiment Results}
We train IFIB-C and baselines on five synthetic datasets and gauge the gap between \(\hat{p}(m, t)\) and \(p(m, t)\) using Spearman coefficient and \(L^1\) distance. The result shows that IFIB-C consistently learns more accurate distributions than all baselines. This conclusion gives us the confidence to apply IFIB-C to real-world datasets. Detailed experiment results are available in \cref{app:synthetic_cate_results}.

Generally, the real-world data possess complicated temporal patterns and intricate correlations between marks which challenges the marked TPP modeling. This section trains IFIB-C and baselines on the four real-world datasets. They are evaluated by comparing the performances on two prediction problems discussed in \cref{sec:application}, i.e., time-event (\cref{tab:IFIB_real_world_mae} and \cref{tab:IFIB_real_world_f1}) and event-time (\cref{tab:IFIB_real_world_mae_e} \cref{tab:IFIB_real_world_f1-e}). The numbers in bold or underlined indicate the best or the second-best value. We only report the time related performance of baseline RMTPP and LogNormMix in the time-event task because these models cannot determine the mark based on history \(\mathcal{H}_{t_l}\) and time \(t\). Meanwhile, as THP always returns meaningless outputs on Retweet, MOOC, and Bookorder, no result of THP on these datasets is reported.

\subsubsection{Event-Time Prediction Problem}
\label{sec:mtpp_et_performance}

\begin{table*}[!ht]
    \caption{Event-time prediction problem on real-world datasets measured by MAE-E$@x\%$(\(25\%, 50\%, 75\%\)).}
    \centering
    \begin{small}
\vskip 0.15in
    \begin{tabular}{lccccc}
        \toprule
        & IFIB(Ours) & FENN & FullyNN & SAHP & THP \\
        \midrule
        \multirow{3}{*}{Retweet} & \textbf{4.6142\tiny{\(\pm\)0.0328}} & \underline{10.018\tiny{\(\pm\)4.0837}} & 16.092\tiny{\(\pm\)7.5513} & 16.970\tiny{\(\pm\)0.1569} & /	\\
        & \textbf{28.384\tiny{\(\pm\)0.0825}} & \underline{50.755\tiny{\(\pm\)15.025}} & 70.959\tiny{\(\pm\)28.580} & 116.05\tiny{\(\pm\)5.4806} & / \\ 
        & \textbf{232.96\tiny{\(\pm\)1.0114}} & \underline{354.22\tiny{\(\pm\)51.368}} & 367.64\tiny{\(\pm\)74.523} & 636.25\tiny{\(\pm\)3.3811} & / \\
        \midrule
        \multirow{3}{*}{SO} & \textbf{0.1179\tiny{\(\pm\)0.0004}} & 0.1672\tiny{\(\pm\)0.0066} & 0.1530\tiny{\(\pm\)0.0041} & \underline{0.1377\tiny{\(\pm\)0.0039}} & 0.1535\tiny{\(\pm\)0.0004} \\
        & \textbf{0.2821\tiny{\(\pm\)0.0009}} & 0.4408\tiny{\(\pm\)0.0121} & 0.3519\tiny{\(\pm\)0.0052} & \underline{0.3327\tiny{\(\pm\)0.0075}} & 0.3378\tiny{\(\pm\)0.0004} \\ 
        & 0.7597\tiny{\(\pm\)0.0051} & 0.9103\tiny{\(\pm\)0.0177} & \textbf{0.6717\tiny{\(\pm\)0.0148}} & \underline{0.6795\tiny{\(\pm\)0.0154}} & 0.5531\tiny{\(\pm\)0.1525} \\
        \midrule
        \multirow{3}{*}{MOOC} &  \textbf{4.2347\tiny{\(\pm\)0.1265}}  & 1162.9\tiny{\(\pm\)318.92} & 2704.8\tiny{\(\pm\)190.60} &  3380.8\tiny{\(\pm\)1782.2}  &  / \\
        &  \textbf{25.504\tiny{\(\pm\)0.2938}}  &  4157.1\tiny{\(\pm\)29.943}  & 4277.2\tiny{\(\pm\)6.8777} & \(>\) 500,000 &   / \\ 
        &  \textbf{283.38\tiny{\(\pm\)2.9122}}  & 4657.1\tiny{\(\pm\)11.563} &  4653.4\tiny{\(\pm\)14.971}   &  \(>\) 500,000 &  / \\
        \midrule
        \multirow{3}{*}{BO} & \textbf{0.0210\tiny{\(\pm\)0.0007}} & 200.53\tiny{\(\pm\)0.1841} & 200.17\tiny{\(\pm\)1.3407} & \underline{0.0220\tiny{\(\pm\)0.0019}} & / \\
        & \textbf{0.0298\tiny{\(\pm\)0.0007}} & 203.66\tiny{\(\pm\)0.0000} & 203.66\tiny{\(\pm\)0.0000} & \underline{0.0469\tiny{\(\pm\)0.0004}} & / \\ 
        & \textbf{0.2092\tiny{\(\pm\)0.0008}} & 203.71\tiny{\(\pm\)0.0000} & 203.71\tiny{\(\pm\)0.0000} & \underline{0.2722\tiny{\(\pm\)0.0184}} & / \\
        \bottomrule
    \end{tabular}
    \end{small}
    \label{tab:IFIB_real_world_mae_e}
\vskip -0.1in
\end{table*}

In this test, we process each mark $m$ if its probability to be the next event is non-zero. On the condition that the mark of the next event is $m$, the time of the next event is predicted. For the real mark of the next event, the difference between the prediction and real-time is used to measure the performance, i.e., MAE-E. The results are in \cref{tab:IFIB_real_world_mae_e}. 

\begin{table}[!ht]
    \caption{Event-time prediction problem on real-world datasets measured by macro-F1.}
    \vskip 0.15in
    \centering
    \begin{small}
    \begin{tabular}{llcccc}
        \toprule
                                     & Retweet & SO & MOOC & BO \\
        \midrule
        IFIB-C(Ours)                & \underline{0.3576\tiny{\(\pm\)0.0008}} & 0.1085\tiny{\(\pm\)0.0009} & \textbf{0.3684\tiny{\(\pm\)0.0019}} & \textbf{0.6021\tiny{\(\pm\)0.0007}} \\
        FullyNN                   & 0.2316\tiny{\(\pm\)0.0000} & 0.0121\tiny{\(\pm\)0.0000} & 0.0005\tiny{\(\pm\)0.0000} & 0.3339\tiny{\(\pm\)0.0000} \\
        % MFullyNN                     & 0.3616 & 0.1400 & 0.0000 & 0.5354 \\
        FENN                      & \textbf{0.3646\tiny{\(\pm\)0.0010}} & 0.0930\tiny{\(\pm\)0.0020} & 0.1698\tiny{\(\pm\)0.0042} & 0.3902\tiny{\(\pm\)0.0450} \\
        % \toprule[0.05em]
        % LogNormMix                   &    \   &    \   &    \   &    \   \\
        %\midrule
        SAHP                      & 0.3558\tiny{\(\pm\)0.0021} & \underline{0.1219\tiny{\(\pm\)0.0050}} & \underline{0.2572\tiny{\(\pm\)0.0312}} & \underline{0.5980\tiny{\(\pm\)0.0009}} \\
        THP                       & / & \textbf{0.1373\tiny{\(\pm\)0.0025}} & / & / \\
        \bottomrule
    \end{tabular}
    \end{small}
    \label{tab:IFIB_real_world_f1-e}
\vskip -0.1in
\end{table}

We can observe IFIB-C demonstrates superiority over baselines on all datasets. Compared with the results for the time-event prediction problem, the relative advantage of IFIB-C is more significant for the event-time prediction problem. As discussed in \cref{sec:application}, the time prediction for a given mark is derived from \cref{eqn:pred_time}, which depends on the joint PDF $p^*(m,t)$. The well-suited $p^*(m,t)$ leads to a better time prediction. The results in \cref{tab:IFIB_real_world_mae_e} implicate that IFIB-C can model $p^*(m,t)$ in a better way than baselines. Further analysis of results in \cref{tab:IFIB_real_world_mae_e} are available in \cref{app:mae_e_analysis}.

In addition, the mark of the next event is predicted following \cref{eqn:pred_time1}. The performances of IFIB-C and baselines measured in macro-F1 are reported in \cref{tab:IFIB_real_world_f1-e}. IFIB-C retains its advantage over other integral-based methods. In summary, the event-time prediction performance further affirms that \(\Gamma^*(m, t)\) in IFIB-C is a better estimation target than \(\Lambda^*(t)\) in FENN.

\subsubsection{Time-event Prediction Problem}
\label{sec:mtpp_te_performance}
\begin{table*}[!ht]
    \caption{Time-event prediction performance on real-world datasets measured by MAE$@x\%$(\(25\%\), \(50\%\), \(75\%\)).}
    \begin{small}
    \centering
    \vskip 0.15in
    \begin{tabular}{lccccccc}
        \toprule
        & IFIB-C(Ours) & FENN & FullyNN & RMTPP & LogNormMix & SAHP & THP \\
        \midrule
        \multirow{3}{*}{\rotatebox[origin=c]{90}{Retweet}} & \textbf{4.6309\tiny{\(\pm\)0.0422}} & \underline{4.6344\tiny{\(\pm\)0.0350}} & 4.6957\tiny{\(\pm\)0.0311} & 4.8436\tiny{\(\pm\)0.0161} & 5.7855\tiny{\(\pm\)0.7749} &  4.6991\tiny{\(\pm\)0.0934} & / \\
        & 28.545\tiny{\(\pm\)0.0839} & 29.136\tiny{\(\pm\)0.0352} & 28.575\tiny{\(\pm\)0.1942} & 30.600\tiny{\(\pm\)0.0134} & \underline{28.380\tiny{\(\pm\)1.0108}} & \textbf{28.037\tiny{\(\pm\)0.4119}} & / \\ 
        & \underline{238.91\tiny{\(\pm\)0.5870}} & 255.62\tiny{\(\pm\)2.5578} & \textbf{238.75\tiny{\(\pm\)0.8049}} & 269.90\tiny{\(\pm\)0.9863} & 249.68\tiny{\(\pm\)2.7573} & 261.44\tiny{\(\pm\)7.6941} & / \\ \midrule
        \multirow{3}{*}{\rotatebox[origin=c]{90}{SO}} & \textbf{0.1411\tiny{\(\pm\)0.0010}} & 0.1972\tiny{\(\pm\)0.0009} & \underline{0.1443\tiny{\(\pm\)0.0011}} & 0.1477\tiny{\(\pm\)0.0026} & 0.1588\tiny{\(\pm\)0.0128} &  0.1463\tiny{\(\pm\)0.0014} & 0.1546\tiny{\(\pm\)0.0004} \\
        & \textbf{0.3347\tiny{\(\pm\)0.0004}} & 0.4287\tiny{\(\pm\)0.0007} & \underline{0.3352\tiny{\(\pm\)0.0011}} & 0.3397\tiny{\(\pm\)0.0060} & 0.3624\tiny{\(\pm\)0.0375} & 0.3422\tiny{\(\pm\)0.0017} & 0.3394\tiny{\(\pm\)0.0014} \\ 
         & 0.6644\tiny{\(\pm\)0.0043} & 0.6867\tiny{\(\pm\)0.0017} & \textbf{0.6527\tiny{\(\pm\)0.0013}} & 0.6675\tiny{\(\pm\)0.0069} & 0.7818\tiny{\(\pm\)0.0464} & 0.6660\tiny{\(\pm\)0.0007} & \underline{0.6621\tiny{\(\pm\)0.0005}} \\
        \midrule    
        \multirow{3}{*}{\rotatebox[origin=c]{90}{MOOC}} & \underline{6.5689\tiny{\(\pm\)0.0996}} & 34.820\tiny{\(\pm\)2.4208} & \textbf{5.8824\tiny{\(\pm\)0.1621}} & 19.217\tiny{\(\pm\)0.9800} & 8.4106\tiny{\(\pm\)0.9752} & 7.8945\tiny{\(\pm\)1.1120} & / \\
        & \textbf{38.795\tiny{\(\pm\)1.4440}} & 214.74\tiny{\(\pm\)14.587} & 47.454\tiny{\(\pm\)0.2333} & 147.15\tiny{\(\pm\)5.6299} & \underline{39.622\tiny{\(\pm\)7.1465}} & 135.49\tiny{\(\pm\)17.609} & / \\ 
        & \textbf{368.25\tiny{\(\pm\)0.8369}} & 825.98\tiny{\(\pm\)24.410} & 386.98\tiny{\(\pm\)4.8503} & 853.24\tiny{\(\pm\)29.962} & \underline{412.06\tiny{\(\pm\)11.384}} & 1262.1\tiny{\(\pm\)273.55} & / \\
        \midrule
        \multirow{3}{*}{\rotatebox[origin=c]{90}{BO}} & 0.0213\tiny{\(\pm\)0.0008} & \textbf{0.0138\tiny{\(\pm\)0.0005}} & \textbf{0.0138\tiny{\(\pm\)0.0003}} & 0.0603\tiny{\(\pm\)0.0053} & \underline{0.0167\tiny{\(\pm\)0.0000}} & 0.0190\tiny{\(\pm\)0.0017} & / \\
        & \textbf{0.0297\tiny{\(\pm\)0.0007}} & 0.0415\tiny{\(\pm\)0.0006} & 0.0413\tiny{\(\pm\)0.0003} & 0.1425\tiny{\(\pm\)0.0020} & \underline{0.0333\tiny{\(\pm\)0.0000}} & 0.0444\tiny{\(\pm\)0.0010} & / \\ 
        & \textbf{0.2091\tiny{\(\pm\)0.0004}} & 0.2391\tiny{\(\pm\)0.0186} & 0.2488\tiny{\(\pm\)0.0118} & 0.5189\tiny{\(\pm\)0.0128} & \underline{0.2112\tiny{\(\pm\)0.0079}} & 0.2353\tiny{\(\pm\)0.0256} & / \\
        \bottomrule
    \end{tabular}
    \end{small}
    \label{tab:IFIB_real_world_mae}
\vskip -0.1in
\end{table*}

\begin{table}[!ht]
    \caption{Time-event prediction performance on real-world datasets measured by macro-F1.}
    \centering
    \begin{small}
\vskip 0.15in
    \begin{tabular}{lcccc}
        \toprule
                                  & Retweet & SO & MOOC & BO \\
        \midrule
        IFIB-C(Ours)              & \underline{0.3530\tiny{\(\pm\)0.0016}} & 0.0797\tiny{\(\pm\)0.0020} & \textbf{0.3499\tiny{\(\pm\)0.0064}} & \underline{0.6002\tiny{\(\pm\)0.0016}} \\
        FENN                      & 0.3468\tiny{\(\pm\)0.0010} & 0.0147\tiny{\(\pm\)0.0020} & 0.0951\tiny{\(\pm\)0.0054} & 0.4006\tiny{\(\pm\)0.0530} \\
        FullyNN                   & 0.2315\tiny{\(\pm\)0.0000} & 0.0121\tiny{\(\pm\)0.0000} & 0.0005\tiny{\(\pm\)0.0000} & 0.3339\tiny{\(\pm\)0.0000} \\
        % \toprule[0.05em]
        % LogNormMix                   &    \   &    \   &    \   &    \   \ \
        % RMTPP                     & 0.3552\tiny{\(\pm\)0.0003} & 0.0779\tiny{\(\pm\)0.0061} & 0.0870\tiny{\(\pm\)0.0090} & \underline{0.6014\tiny{\(\pm\)0.0007}} \\
        % LogNormMix                & \textbf{0.3735\tiny{\(\pm\)0.0004}} & \textbf{0.1054\tiny{\(\pm\)0.0015}} & \underline{0.2894\tiny{\(\pm\)0.0447}} & 0.6029\tiny{\(\pm\)0.0004} \\
        %\midrule
        SAHP                      & \textbf{0.3544\tiny{\(\pm\)0.0020}} & \underline{0.1185\tiny{\(\pm\)0.0040}} & \underline{0.3340\tiny{\(\pm\)0.0109}} & \textbf{0.6011\tiny{\(\pm\)0.0006}} \\
        THP                       & / & \textbf{0.1380\tiny{\(\pm\)0.0023}} & / & / \\
        \bottomrule
    \end{tabular}
    \end{small}
    \label{tab:IFIB_real_world_f1}
\vskip -0.1in
\end{table}

As in \cref{eqn:pred_t1} and \cref{eqn:pred_t2}, the time-event prediction problem first predicts when the next event will happen, then predicts which mark the next event is most likely to be at the time. The metric for evaluating time prediction is MAE, and the metric for evaluating mark prediction is macro-F1. The test results are in \cref{tab:IFIB_real_world_mae} and \cref{tab:IFIB_real_world_f1}. 

As shown in \cref{tab:IFIB_real_world_mae}, IFIB-C outperforms the integral-based methods (FENN and FullyNN) in general. It means IFIB-C provides more accurate time prediction for more events. Even in situations where IFIB-C does not show the best performance, the performance of IFIB-C is highly comparable. When compared with other baselines, IFIB-C demonstrates more significant advantages. This may be caused by the fact that these baselines are more or less affected by the intensity functions predefined.
% On MOOC, the training process of methods, except IFIB-C, LogNormMix, and RMTPP, completely halts during the training process, i.e., the loss, macro-F1, and MAE$@x\%$ are frozen. Therefore, no test results are reported for them. 
%SAHP and THP demonstrate better performance on MOOC.
% We believe all these are due to a large number of marks (97) and the relatively short event sequences (57 events on average) in MOOC. It makes capturing the temporal correlation between events highly challenging. 
% The better performance of SAHP and THP may attribute to the transformer applied.

In \cref{tab:IFIB_real_world_f1}, IFIB-C beats all baselines in general in terms of macro-F1. It indicates IFIB-C is competent to time-event tasks. % like other baselines.
%accurately predict the mark of the next event compared to other methods. %IFIB marginally outperforms FullyNN because FullyNN could not distinguish \(p^*(m, t)\) because of computation graph overlap. However, we find that  
Looking closely, IFIB-C consistently defeats FENN. Considering the main difference between IFIB-C and FENN is that IFIB-C outputs \(\Gamma^*(m, t)\) and FENN outputs \(\Lambda^*(m, t)\), the test results provide evidence that \(\Gamma^*(m, t)\) can be better estimated than \(\Lambda^*(m, t)\). %Moreover, SAHP and THP, equipped with transformer-powered historical encoders, receive higher macro-F1 than FullyNN and FENN, but still, get defeated by IFIB which utilizes an RNN for history embedding. %Transformers have been proven to provide more powerful history embeddings than RNNs. It could be interesting to see IFIB's performance with Transformer-based historical encoders in future work.

\section{Experiments on IFIB-N}
We evaluated IFIB-N on five synthetic datasets, including Hawkes\_1, Hawkes\_2, Poisson, Self-correct, and Stationary Renewal \cite{omi_fully_2019}, and three real-world datasets including Earthquake, Citibike, and COVID-19 \cite{chen_neural_2020}. Detailed information about these datasets is available in \cref{app:datasets}. Our investigation did not identify proper baselines as most existing studies related to joint distribution $p^*(m,t)$ focus on categorical marks.    

Since the mark is continuous, we cannot use macro-F1 to measure the mark prediction performance. Instead, we use DV (\underline{D}istance between the predicted \underline{V}ector and the ground truth) in pair with MAE and MAE-E mentioned in \cref{sec:em}. We report Q1, Q2, and Q3 of DV, MAE, and MAE-E for reliable measurement. For the synthetic datasets, we use the same metrics, i.e., Spearman coefficient, \(L^1\) distance, and relative NLL loss, to demonstrate that IFIB-N learns the true distribution \(p^*(m, t)\) with high fidelity. Detailed information about all mentioned metrics is available in \cref{app:metric}, and we present all experiment results on synthetic datasets in \cref{app:synthetic_num_results}.

In Table 5 and \cref{tab:cifib_real_mae_e}, the performance of IFIB-N on three real-world datasets, Citibike, COVID-19, and Earthquake is reported. IFIB-N is the first marked TPP model that explicitly enforces normalized \(p^*(\mathbf{m}, t)\) and \(P^*(\mathcal{C}(\mathbf{m}), t)\) for marks in a multi-dimensional continuous space. %It calculates the integral \(\Gamma^*(\mathbf{m}, t)\) over time and the mark space, then obtaining the \((n+1)\)-rank gradient of \(\Gamma^*(\mathbf{m}, t)\) as \(p(t, m_1, m_2, \cdots, m_n)\). By doing so,  finally successfully elude the computation-impossible multi-dimensional numerical integration methods. 
The results prove that IFIB-N could properly model \(P^*(t) = \int_{t_l}^{t}{\int_{m\in \mathbf{M}}{p^*(\mathbf{m}, \tau)d\mathbf{m}d\tau}}\) and \(p^*(\mathbf{m}, t)\) for the time-event prediction task, and properly model \(p^*(m) = \int_{t_l}^{+\infty}{p^*(\mathbf{m}, \tau)d\tau}\) and \(P^*(\mathcal{C}(\mathbf{m}), t)\) for the event-time prediction task.

\begin{table}[ht]
    \centering
    \begin{small}
    \caption{Time-event prediction performance of IFIB-N on real-world datasets}
    \vskip 0.15in
    \begin{tabular}{llccc}
        \toprule
                             &          & Citibike & COVID-19 & Earthquake \\
        \midrule
        \multirow{3}{*}{MAE} & MAE\(@25\%\) & 0.0293\tiny{\(\pm\)0.0000} & 0.0102\tiny{\(\pm\)0.0001} & 0.0650\tiny{\(\pm\)0.0015} \\
                             & MAE\(@50\%\) & 0.0607\tiny{\(\pm\)0.0000} & 0.0216\tiny{\(\pm\)0.0000} & 0.1646\tiny{\(\pm\)0.0007} \\
                             & MAE\(@75\%\) & 0.1299\tiny{\(\pm\)0.0001} & 0.0560\tiny{\(\pm\)0.0004} & 0.3108\tiny{\(\pm\)0.0007} \\
        \midrule           
        \multirow{3}{*}{DV}  & DV\(@25\%\) & 0.5176\tiny{\(\pm\)0.0003} & 3.5627\tiny{\(\pm\)0.0361} & 3.0615\tiny{\(\pm\)0.0684} \\
                             & DV\(@50\%\) & 0.5357\tiny{\(\pm\)0.0002} & 3.7334\tiny{\(\pm\)0.0205} & 6.3055\tiny{\(\pm\)0.1422} \\
                             & DV\(@75\%\) & 0.5515\tiny{\(\pm\)0.0004} & 3.9146\tiny{\(\pm\)0.0276} & 10.379\tiny{\(\pm\)0.2032} \\
        \bottomrule
    \end{tabular}
    \end{small}
    \vskip -0.1in
    \label{tab:cifib_real_mae_a}
\end{table}

\begin{table}[ht]
    \centering
    \caption{Event-time prediction performance of IFIB-N on real-world datasets}
    \vskip 0.15in
    \begin{small}
    \begin{tabular}{llccc}
        \toprule
                               &          & Citibike & COVID-19 & Earthquake \\
        \midrule
        \multirow{3}{*}{MAE-E} & MAE\(@25\%\) & 0.0293\tiny{\(\pm\)0.0000} & 0.0102\tiny{\(\pm\)0.0001} & 0.0650\tiny{\(\pm\)0.0015} \\
                             & MAE\(@50\%\) & 0.0607\tiny{\(\pm\)0.0000} & 0.0216\tiny{\(\pm\)0.0000} & 0.1646\tiny{\(\pm\)0.0007} \\
                             & MAE\(@75\%\) & 0.1299\tiny{\(\pm\)0.0001} & 0.0560\tiny{\(\pm\)0.0004} & 0.3108\tiny{\(\pm\)0.0007} \\
        \midrule           
        \multirow{3}{*}{DV}  & DV\(@25\%\) & 0.5176\tiny{\(\pm\)0.0003} & 3.5627\tiny{\(\pm\)0.0361} & 3.0615\tiny{\(\pm\)0.0684} \\
                             & DV\(@50\%\) & 0.5357\tiny{\(\pm\)0.0002} & 3.7334\tiny{\(\pm\)0.0205} & 6.3055\tiny{\(\pm\)0.1422} \\
                             & DV\(@75\%\) & 0.5515\tiny{\(\pm\)0.0004} & 3.9146\tiny{\(\pm\)0.0276} & 10.379\tiny{\(\pm\)0.2032} \\
        \bottomrule
    \end{tabular}
    \end{small}
    \vskip -0.1in
    \label{tab:cifib_real_mae_e}
\end{table}

\section{Conclusion}
This paper proposed a framework IFIB to model joint conditional PDF $p^*(\mathtt{m},t)$ by exploring the relationship between $p^*(\mathtt{m},t)$ and its marginal probability of mark, i.e., $p^*(\mathtt{m})$, with neural networks. The advantages of IFIB have been demonstrated in three aspects. First, since no intensity function is involved, it remarkably simplifies the process to compel the essential mathematical restrictions. Second, it can be naturally applied to model TPP with categorical marks as well as numeric marks in a multi-dimensional continuous space. Third, it supports different applications regarding the time and mark prediction of the next event. In the future study, we plan to extend IFIB to model marked TPP problems that encounter the complex mark with both categorical information and numeric values in a multi-dimensional continuous space, for example, the observations of different species in a spatial region over time.

\clearpage
\bibliography{reference.bib}
% \bibliographystyle{style/icml2023}

%
% The appendix starts here.
%
\clearpage
\appendix

\section{The Conditional Joint PDF}
\label{appendix:mtpp_pdf}
Without loss of generality, we assume the mark is categorical. For mark \(m\), we define a conditional intensity function \(\lambda^*(m, t)\):
\begin{equation}
\begin{aligned}
\label{eqn:lambda_i}
    & \lambda^*(m = k_i, t) = \lambda(m = k_i, t|\mathcal{H}_t) \\
    &= \lim_{\Delta t \rightarrow 0}{\frac{P(m = k_i, t \in [t, t + \Delta t)|\mathcal{H}_{t-})}{\Delta t}} \\
    &= \lim_{\Delta t \rightarrow 0}{\frac{p(m = k_i, t \in [t, t + \Delta t)|\mathcal{H}_{t_l}) \Delta t}{P(\forall j \in \mathbb{N}^+, t_j \notin (t_l, t)|\mathcal{H}_{t_l}) \Delta t}} \\
    &= \lim_{\Delta t \rightarrow 0}{\frac{p(m = k_i, t \in [t, t + \Delta t)|\mathcal{H}_{t_l})}{P(\forall j \in \mathbb{N}^+, t_j \notin (t_l, t)|\mathcal{H}_{t_l})}} \\
    &= \frac{p(m = k_i, t \in [t, t + dt)|\mathcal{H}_{t_l})}{P(\forall j \in \mathbb{N}^+, t_j \notin (t_l, t)|\mathcal{H}_{t_l})}
\end{aligned}
\end{equation}
where \(\mathcal{H}_{t_l}\) is the history up to (including) the most recent event, \(\mathcal{H}_{t-}\) is the history up to (excluding) the current time, \(P(\forall j \in \mathbb{N}^+, t_j \notin (t_l, t)|\mathcal{H}_{t_l})\) represents the probability that no event is observed in time interval \((t_l, t)\) given \(\mathcal{H}_{t_l}\).

%Now, we need to solve the conditional probability distribution \(p\) in \cref{eqn:lambda_i} with provided \(\lambda^*(m, t)\). 

We denote \(P^{'}_m((t_1,t_2)|\mathcal{H}_{t_l})\) for the conditional probability that an event \(m\) happens in $(t_1,t_2)$. %, and \(p^{'}_m(t|\mathcal{H}_{t_l})\) for the conditional probability density distribution of event \(m\) at time $t>t_l$. 
Following the definition of simple TPP that at most one event happens at every timestamp \(t\), the probability that no event occurs in \((t_l, t)\) is:
\begin{equation}
\begin{aligned}\label{eqn:noevent}
& P(\forall j \in \mathbb{N}^+, t_j \notin (t_l, t)|\mathcal{H}_{t_l}) \\ 
&= 1 - \sum_{w\in \mathrm{M}}{P^{'}_w((t_l,t)|\mathcal{H}_{t_l})\prod_{n\in \mathrm{M}, n \neq w}({1 - P^{'}_n((t_l,t)|\mathcal{H}_{t_l})})} \\
&= 1 - \sum_{w\in \mathrm{M}}{\frac{P^{'}_w((t_l,t)|\mathcal{H}_{t_l})}{1-P^{'}_w((t_l,t)|\mathcal{H}_{t_l})}}\prod_{n\in M}({1-P^{'}_n((t_l,t)|\mathcal{H}_{t_l})}) \\
&= 1 - \sum_{w\in \mathrm{M}}{P(w,t|\mathcal{H}_{t_l})} = 1 - \sum_{w\in \mathrm{M}}{P^*(w,t)} 
\end{aligned}
\end{equation}
where
\begin{equation}\label{eqn:P_mt}
P^*(w,t) = \frac{P^{'}_w((t_l,t)|\mathcal{H}_{t_l})}{1-P^{'}_w((t_l,t)|\mathcal{H}_{t_l})}\prod_{n\in \mathrm{M}}({1 - P^{'}_n((t_l, t)|\mathcal{H}_{t_l}))}
\end{equation}
The conditional joint PDF that the next event is $m$ and occurs in $[t, t+dt)$ is:
\begin{subequations}
\begin{align}\label{eqn:pnext}
    & p(m = k_i, t \in [t, t + \Delta t)|\mathcal{H}_{t_l}) = \frac{dP^*(m = k_i,t)}{dt} \\ 
    & \int_{t_l}^{t}{p(m = k_i, t \in [t, t + \Delta t)|\mathcal{H}_{t_l})d\tau} = P^*(m = k_i, t)
\end{align}
\end{subequations}

% Now we could obtain \(\lambda^*(m, t)\):
% \begin{equation}
% \begin{aligned}
% \label{eqn:intensity_i_source}
%     &\lambda^*(m, t) \\
%     &= \frac{p(next\ event\ is\ m\ in\ [t, t + dt)|\mathcal{H}_{t_l})}{P(no\ event\ in\ (t_l, t]|\mathcal{H}_{t_l})}\\
%     &= \frac{d ((\prod_{n\in M}{1 - P^{'}_n(t|\mathcal{H}_{t_l}))}\frac{P^{'}_m(t|\mathcal{H}_{t_l})}{1-P^{'}_m(t|\mathcal{H}_{t_l})})/ dt}{1 - (\prod_{n\in M}{1 - P^{'}_n(t|\mathcal{H}_{t_l}))}\sum_{w\in M}{\frac{P^{'}_w(t|\mathcal{H}_{t_l})}{1 - P^{'}_w(t|\mathcal{H}_{t_l})}}}
% \end{aligned}
% \end{equation}

In this study, $p^*(m,t)$, shorthand of \(p(m, t|\mathcal{H}_{t_l})\), is the formal representation of $p(m = k_i, t \in [t, t + \Delta t)|\mathcal{H}_{t_l})$. Note $P^*(m, t)$ in \cref{eqn:P_mt} is the probability that only one event happens in interval \([t, t + dt)\) and the mark is $m$. This ensures the marked TPP represented by $p^*(m,t)$ is simple. By integrating \cref{eqn:pnext} and \cref{eqn:noevent} in \cref{eqn:lambda_i}, we have  
% \begin{equation}
%   \lambda^*(m, t) = \frac{p^*(m, t)}{1 - \sum_{w\in M}{P(w,t|\mathcal{H}_{t_l})}}  
% \end{equation}
% We have
\begin{equation}
  p^*(m, t) = \lambda^*(m, t)(1 - \sum_{w\in \mathrm{M}}{P^*(w,t)})  
\end{equation}
where $\sum_{w\in \mathrm{M}}{P^*(w,t)})$ is calculated from the sum of \cref{eqn:lambda_i} over marker \(m\):
\begin{equation}
\label{eqn:mtpp_sum_P}
    \sum_{w\in \mathrm{M}}{P^*(w, t)} = 1 - \exp(-\int_{t_l}^{t}{\sum_{n\in \mathrm{M}}{\lambda^*(n, \tau)}d\tau})
\end{equation}
Then, we solve $p^*(m, t)$:
\begin{equation}
p^*(m, t) = \lambda^*(m, t)\exp(-\int_{t_l}^{t}{\sum_{n\in \mathrm{M}}{\lambda^*(n, \tau)}d\tau})
\end{equation}
which is equivalent with \cref{eqn:mtpp}. We won't go into details about numeric marks but directly provide the expression of joint PDF \(p(\mathbf{m}, t)\) in \cref{eqn:pdf_continuous_m}. One could replace the summation of all categorical marks by the integration over the multi-dimensional continuous space to smoothly adapt the conclusion for categorical marks to the numeric marks.

\begin{equation}
\label{eqn:pdf_continuous_m}
p^*(\mathbf{m}, t) = \lambda^*(\mathbf{m}, t)\exp(-\int_{t_l}^{t}{\int_{\mathbf{n} \in \mathbf{M}}{\lambda^*(\mathbf{n}, \tau)}d\tau d\mathbf{n}})
\end{equation}

\section{Analysis on FullyNN and FENN}
The analysis is based on categorical marks only since FullyNN and FENN cannot handle numeric marks in multi-dimensional continuous space. 

\subsection{Why FullyNN incapable of marked TPP}
\label{appendix:FullyNN}
FullyNN sets the learning target of NNs to \(\Lambda^*(t)\), the integral of intensity functions \(\lambda^*(t)\). This idea works when no event mark information is present. However, when we require FullyNN to learn different intensity functions for different event marks, the limitation emerges as FullyNN cannot allocate different computation graphs for different marks.

To understand this, we should recognize how FullyNN calculates the intensity function. Following the definition in \cite{omi_fully_2019}, a FullyNN can be written in the following expression:
\begin{equation}
\label{eqn:FullyNN}
    \Lambda^*(t) 
    % &= \operatorname{softplus}(\mathbf{w}_0^{\top}\tanh(\mathbf{W}_2(\tanh(\mathbf{W}_1(\mathbf{h} + \mathbf{f}(t)) + \mathbf{b_1}) + \mathbf{b}_2) + b_0) \\
    = \operatorname{FullyNN}(t, \mathbf{h}) = \operatorname{IEM}(\mathbf{f}(t), \mathbf{h})
\end{equation}
where \(\mathbf{f}(t)\) represents a monotonic-increasing function mapping the time \(t\) into a vector, \(\mathbf{h}\) is the history embedding, and \(\operatorname{IEM}\) refers to the integral estimation module. From \cref{eqn:FullyNN}, we could derive the intensity function as:
\begin{equation}
\begin{aligned}
\label{eqn:FullyNN_def}
    \lambda^*(t) = \frac{\partial \Lambda^*(t)}{\partial t} = \frac{\partial \operatorname{IEM}(\mathbf{f}(t), \mathbf{h})}{\partial \mathbf{f}(t)}\frac{\partial \mathbf{f}(t)}{\partial t}
\end{aligned}
\end{equation}

% \begin{equation}
% \begin{aligned}
% \label{eqn:FullyNN_def}
%     \lambda^*(t) = \frac{\partial \Lambda^*(t)}{\partial t} \\ &= \frac{\partial \operatorname{FullyNN}(t, \mathbf{h})}{\partial t} = \frac{\partial \operatorname{IEM}(\mathbf{f}(t), \mathbf{h})}{\partial \mathbf{f}(t)}\frac{\partial \mathbf{f}(t)}{\partial t}
% \end{aligned}
% \end{equation}
If one generalizes the FullyNN from mark-agnostic to mark-aware by simply expanding the input time from \(t\) to \(\mathbf{t} = [t, t, t, \cdots, t]^{\top}\), each for one of the $|\mathrm{M}|$ marks, and they share the same vector \(\mathbf{v}\) for generating the same \(\mathbf{f}(t)\)s as input of IEM. By letting the corresponding intensity integral be \(\mathbf{\Lambda}^*(\mathbf{t}) = [\Lambda^*(m = k_1, t), \Lambda^*(m = k_2, t), \Lambda^*(m = k_3, t), \cdots, \Lambda^*(m = k_{|\mathrm{M}|}, t)]^{\top}\), we could find the Jacobian Matrix \(D_{\mathbf{t}}\mathbf{\Lambda}^*(t)\) is:
\begin{equation}
\begin{aligned}
    & D_{\mathbf{t}}\mathbf{\Lambda}^*(\mathbf{t}) \\ 
    &= \frac{\partial [\Lambda^*(m = k_1, t), \Lambda^*(m = k_2, t), \cdots, \Lambda^*(m = k_{|M|}, t)]^{\top}}{\partial \mathbf{t}} \\
    &= 
    \begin{pmatrix}
    \frac{\partial \operatorname{IEM(\mathbf{f}(t), \mathbf{h})}}{\partial \mathbf{f}(t)}\mathbf{v} & 0 & \cdots & 0 \\
    0 & \frac{\partial \operatorname{IEM(\mathbf{f}(t), \mathbf{h})}}{\partial \mathbf{f}(t)}\mathbf{v} & \cdots & 0 \\
    \vdots & \vdots & \ddots & \vdots \\
    0 & 0 & \cdots & \frac{\partial \operatorname{IEM(\mathbf{f}(t), \mathbf{h})}}{\partial \mathbf{f}(t)}\mathbf{v} \\
    \end{pmatrix}
\end{aligned}
\end{equation}
which implies that the intensity functions for different marks receive identical distributions, and the event prediction performance would be stuck at \(\frac{1}{|\mathrm{M}|}\). We believe this might explain the shockingly bad event prediction performance in Enguehard et al.'s paper \cite{enguehard_neural_2020}. 
%So, we introduce $\mathbf{v}_{k_1},\cdots,\mathbf{v}_{k_{|M|}}$ to generate $\mathbf{f}(m = k_1, t),\cdots,\mathbf{f}_(m = k_{|M|}, t)$ in IFIB and FENN to customize the intensity function for each mark. In \cref{fig:heatmap_overlap},  

\begin{figure*}[!ht]
    \centering
    \begin{subfigure}{0.32\textwidth}
        \includegraphics[width=\textwidth]{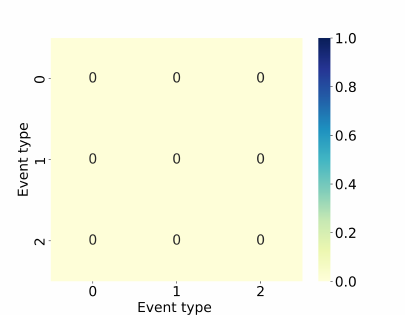}
        \label{fig:heatmap_FullyNN}
        \caption{FullyNN on Retweet dataset.}
    \end{subfigure}
    \begin{subfigure}{0.32\textwidth}
        \includegraphics[width=\textwidth]{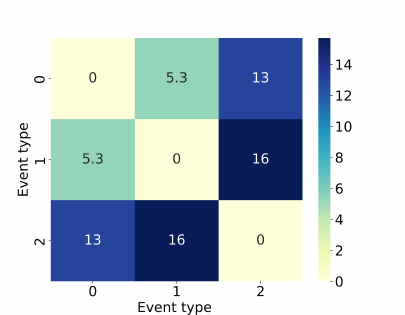}
        \label{fig:heatmap:FullyNN_ete}
        \caption{FENN on Retweet dataset.}
    \end{subfigure}
    \begin{subfigure}{0.32\textwidth}
        \includegraphics[width=\textwidth]{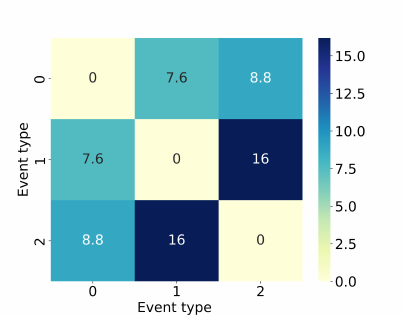}
        \label{fig:heatmap:mFullyNN}
        \caption{IFIB-C on Retweet dataset.}
    \end{subfigure}
    \caption{The $L^1$ distance between distribution \(p^*(m_i,t)\) and distribution \(p^*(m_j, t)\), for each pair of marks ($m_i,m_j$) in $\mathrm{M}$, generated by FullyNN, FENN and IFIB-C on an event sequence in Retweet dataset. FullyNN generates the identical distribution for different marks as the \(L^1\) distance between \(p^*(m_i, t)\) and \(p^*(m_j, t)\) is 0 for each pair of marks ($m_i,m_j$) in $\mathrm{M}$. In contrast, FENN and IFIB-C generate different distributions for different marks.}
    \label{fig:heatmap_overlap}
\end{figure*}

%The central reason is that FullyNN cannot represent different intensity functions by a single computation graph \(\mathcal{G}\). Following the definition of FullyNN in \cref{eqn:FullyNN_def}, we notice that FullyNN represents the integral of an intensity function \(\Lambda^*(m, t)\) as a unique computation graph \(\mathcal{G}_m\). Each intensity function \(\lambda^*(m, t)\) derives from the structure and nodes of the computation graph \(\mathcal{G}_m\) as the backpropagation goes through it. \textbf{If the computation graph \(\mathcal{G}_m\) shares part of itself \(\mathcal{G}^{sub}\) with another computation graph \(\mathcal{G}_n\), then we call these two computation graphs \(\mathcal{G}_m\) and \(\mathcal{G}_n\) have overlaps, which means part of the function \(\Lambda^*(m, t)\) and \(\Lambda^*(n, t)\) is identical.} This might potentially harm the performance on MTPP tasks as each intensity function \(\lambda^*(m, t)\) is supposed to be different from others. Back to the plain FullyNN design, the overlapped computation graph \(\mathcal{G}^{sub}\) is \(\mathcal{G}_m\) because there is only one computation graph available, causing \(\Lambda^*(m, t) = \Lambda^*(n, t)\) for every \(t \in (0, T]\) and marker \(m, n \in M\). \par

\begin{figure*}[ht]
    \centering
    \includegraphics[width=0.75\linewidth]{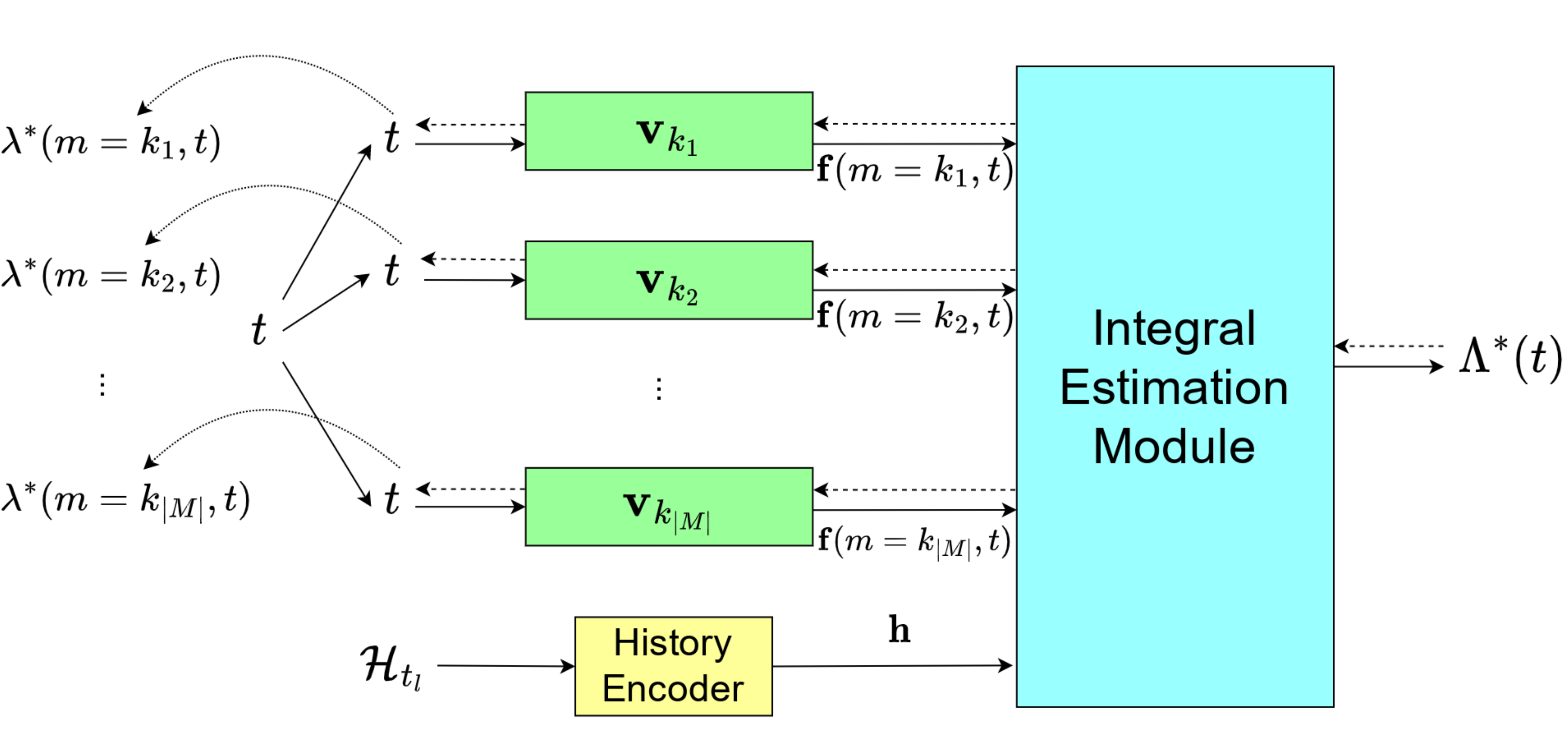}
    \caption{Architecture of Fully Event Neural Network (FENN). The solid arrows refer to forward propagation, the dashed arrows refer to backpropagation, and the curved arrows refer to retrieving the gradient. FENN models \(\Lambda^*(t)\) and obtains \(\lambda^*(m, t)\) by backpropagation.}
    \label{fig:FullyNN_ete}
\end{figure*}

\subsection{Fully Event Neural Network}\label{apx:ETE}
To handle marks in a better way, FullyNN can be extended to FENN (\underline{F}ully \underline{E}vent \underline{N}eural \underline{N}etwork), which models $p^*(m,t)$ based on the conditional intensity $\lambda^*(m,t)$ as defined in \cref{eqn:mtpp}. 
%FENN represents a modified version of FullyNN \cite{omi_fully_2019}, which is originally designed for TPP without the information of the event mark. 
FENN is sketched in \cref{fig:FullyNN_ete}. The history $\mathcal{H}_{t_l}$ is represented as an embedding $\mathbf{h}$ using a LSTM encoder \cite{omi_fully_2019}. FENN needs to model $|\mathrm{M}|$ conditional intensity functions, i.e., $\lambda^*(m,t)$ for all $m\in \mathrm{M}$. The integral of conditional intensity functions across all marks, from the time of the latest event $t_l$ to the current time $t$, is denoted as $\Lambda^*(t)$. The definition of $\Lambda^*(t)$ is given in \cref{eqn:fete_integral}, and the relationship between $\Lambda^*(t)$ and $\lambda^*(m,t)$ is presented in \cref{eqn:fete_intensity}: 
\begin{subequations}
\begin{align}
\begin{split}\label{eqn:fete_integral}
&\Lambda^*(t) = \int_{t_l}^{t}{\sum_{n \in \mathrm{M}}{\lambda^*(n,\tau)}d\tau} = \\ 
& \operatorname{IEM_{FENN}}(\mathbf{f}(m = k_1, t), t),\cdots,\mathbf{f}(m = k_{|\mathrm{M}|}, t), \mathbf{h}), 
\end{split}\\
\label{eqn:fete_intensity}
&\lambda^*(m,t) = \frac{\partial \Lambda^*(t)}{\partial \mathbf{f}(m, t)} \frac{\partial \mathbf{f}(m, t)}{\partial t}, \ \text{for}\ m\in \mathrm{M}.  
\end{align}
\end{subequations}
where $\mathbf{f}(m, t)$ is defined as $\mathbf{v}_{m}(t - t_l)$ for mark $m\in \mathrm{M}$. FENN utilizes $\mathbf{v}_m$ to distinguish event marks to avoid sharing the same computation graph as FullyNN. The integral estimation module (IEM) of FENN contains multiple fully-connected layers with non-negative weights and monotonic-increasing activation functions and ends with an unbounded above softplus function \(\operatorname{softplus}(x) = \log(1 + e^x)\). IEM receives history embedding $\mathbf{h}$ and $\mathbf{f}(m = k_1, t),\mathbf{f}(m = k_2, t),\cdots,\mathbf{f}(m = k_{|\mathrm{M}|}, t)$, outputs $\Lambda^*(t)$. 
The loss function of FENN is the negative logarithm of \(p^*(m, t)\) at every known event \((m_i, t_i)\in \mathcal{S}\), as shown in \cref{eqn:fenn_mtpp_loss}.
\begin{equation}
\begin{aligned}
    L &= \sum_{(m_i, t_i) \in \mathcal{S}}{-\log p^*(m_i, t_i)} \\ &= \sum_{(m_i, t_i) \in \mathcal{S}}{(-\log \lambda^*(m_i, t_i) + \int_{t_l}^{t_i}{\lambda^*(m, \tau)d\tau})}
    \label{eqn:fenn_mtpp_loss}
\end{aligned}
\end{equation}

%Event-oriented MTPP requires the density function $p^*(m)$ for marker $m\in M$, which is the integration of $p^*(m, t)$ over time since the latest event. In turn, $p^*(m, t)$ depends on $\lambda^*(m, t)$ and the integration of $\lambda^*(m, t)$ as in \cref{eqn:mt_m_dist}. Then, the probability $p^*(t|m)$ is estimated with \cref{eqn:mt_t_dist}. 

%Besides the computation graph overlap issue that prevents researchers from correctly adapting II-TPP models to the MTPP tasks,
%jz2: below, receives > inherits
% get it
FENN inherits from FullyNN the capability to instantaneously provide accurate \(\lambda^*(m, t)\) and the integral \(\Lambda^*(t)\), and FENN solves the computation graph overlap issue in FullyNN as evidenced by the test results in \cref{fig:heatmap_overlap}. 
%jz2: below, compel? do not compel > violate the mathematical properties/principle
% done
However, FENN also inherits the weakness of FullyNN: it violates several essential mathematical restrictions \cite{shchur_intensity-free_2020}. To elaborate on this, we first rewrite \cref{eqn:fete_integral} by expanding \(\operatorname{IEM_{FENN}}(\cdot)\).
\begin{subequations}
\begin{align}
    \Lambda^*(t) &= \sum_{m \in M}{\operatorname{softplus}(\Omega^*(m, t) + b)} \\
\label{eqn:fenn_expression}
    \Omega^*(m, t) &= \mathbf{w}^{\top}\operatorname{tanh}(\mathcal{F}_l(\cdots\mathcal{F}_{2}(\mathcal{F}_1([\mathbf{v}_m(t - t_l), \mathbf{h}])))). \\
    \mathcal{F}_i(\mathbf{x}) &= \operatorname{tanh}(\mathbf{W}_{i}\mathbf{x} + \mathbf{b}_i)
\end{align}
\end{subequations}
where \(\mathbf{W}_i\) and \(\mathbf{w}\) are matrices and vectors, respectively, without negative numbers, \(\mathbf{b}_i\) and \(b\) are biases, and \(l\) is the number of non-negative fully-connected layers. With the two expressions above, 
%jz2: below, compel?
we reveal the restrictions that FENN fails to compel:
\begin{enumerate}
    \item The IEM must output 0 when the input is exactly \(t_l\). Otherwise, the model allows future events to occur before the latest historical event, which is unreasonable. Unfortunately, as \(\operatorname{softplus}(x) = \log(1 + e^x)\) is always positive, \(\Lambda^*(t_l)\) would be close to, but never be 0.
    \item The model's output must be unbounded, in other words, $\lim_{t\rightarrow +\infty}\Lambda^*(t) = +\infty$ because the cumulative distribution function, \(P^*(t) = \int_{t_l}^{t}{\sum_{n \in M}p^*(n, \tau)d\tau} =  1 - \exp(-\Lambda^*(t))\), must converge to 1 as \(t \rightarrow +\infty\) if assuming the next event always happens. %{\color{green}(\(P^*(t) = \int_{t}^{+\infty}{p^*(m, \tau)d\tau} =  1 - \exp(-\Lambda^*(t))\), must converge to 1 as \(t\) increases.)} 
    However, similar to FullyNN, because FENN's activation function between the fully-connected layers is \(\tanh(x)\), whose value domain is \((-1, 1)\), the upper bound of \(\Lambda^*(t)\) exists as shown in \cref{eqn:fenn_bounds}, resulting in an unnormalized probability distribution.
    \begin{equation}
    \Lambda^*(t)_{max} = \sum_{m \in M}{\operatorname{softplus}(\mathbf{w}_1^{\top}\mathbf{1} + b_1)} < +\infty
    \label{eqn:fenn_bounds}
    \end{equation}
\end{enumerate}
Moreover, these restrictions are parameter-independent, meaning that one must directly impose them into the model structure, making them more difficult to deal with. In conclusion, although FENN can provide the conditional joint PDF $p^*(m,t)$, its structure is still faulty, which could lead to inferior performance.

%although FENN successfully realises and tackles the computation graph overlap issue by introducing event-wise time embeddings \(\mathbf{v}_m\), its structure is still faulty, which could lead to inferior performance. \par

% Inherited from FullyNN, FENN's structure cannot ensure outputting fixed values at \(t_l\). Instead, it requires a large amount of training data to obtain this knowledge. Meanwhile, FENN limits its output because the activation function, \(tanh(x)\), between each non-negative fully-connected layer is bounded.

% To achieve this, we cannot introduce activation functions whose value range is finite into this structure to avoid any implicit upper bounds. We also need to manage the output value before and after each layer carefully to ensure the integral estimation module always starts from 0. \par

\section{Analysis on LogNormMix}\label{app:lognormmix}
The intensity-free mindset focuses on directly modelling probability distribution while ignoring the intrinsic theory of TPP. LogNormMix \cite{shchur_intensity-free_2020}, powered by this mindset, models the probability of \(p^*(t)\) by a composition of log-normal distributions, whose means and variances derive from history. This approach performs well on synthetic and real-world TPP tasks with closed-form distributions that help it free from intensity estimation. 

These fantastic features propel us to adapt the intensity-free mindset to our task in this paper. Given a categorical mark $m$, the original approach in \cite{shchur_intensity-free_2020} learns a distribution \(p^*(m)\) independent from the time distribution \(p^*(t)\) for event prediction. Such a design removes the connection between mark and time, which may harm the model's performance. 

Therefore, it is intuitive to ask whether we could slightly modify LogNormMix to fit the marked TPP by setting up \(|\mathrm{M}|\) probability distributions for \(|\mathrm{M}|\) categorical markers. Specifically, instead of splitting \(p^*(m, t)\) into two independent distributions, we want to directly extract \(p^*(m, t)\) from the training data like what IFIB-C does. Unfortunately, we cannot do that by following the current mindset of LogNormMix. 

Why? Unlike TPP, the marked TPP concerns the conditional joint PDF \(p^*(m, t)\) where the integral over time and marks should be 1, as shown in \cref{eqn:p_integral_over_time_m}.
\begin{equation}
    \sum_{m \in \mathrm{M}}{\int_{t_l}^{+\infty}{p^*(m, \tau)}d\tau}dm = 1
    \label{eqn:p_integral_over_time_m}
\end{equation}
This equation indicates that \(p^*(m) = \int_{t_l}^{+\infty}{p^*(m, t)}\) for mark $m$ should dwell in \((0, 1)\), and \(\sum_{m\in \mathrm{M}}p^*(m)\) is always 1. However, the mindset of LogNormMix assigns a complete probability distribution \(q_m(t)\) to each mark $m$ where \(q_m(t)\)'s integral from \(t_l\) to infinity is already 1. So \(\sum_{m \in \mathrm{M}}{\int_{t_l}^{+\infty}{q_m(\tau)}d\tau} = |\mathrm{M}|\). This violates \cref{eqn:p_integral_over_time_m}. For the same reason, since \(q_m(t)\) is a complete distribution and independent, whenever we optimise \(q_m(t)\) by minimising its negative logarithm for event \((m_i, t_i)\), \(q_{m_i}(t)\) receives the information to maximize \(q_{m_i}(t_i)\) but the information to update \(q_{m_j}(t)\) is inevitably lost. In conclusion, it is unknown to us how to extend LogNormMix to model $p^*(m,t)$ problem without significant effort. 

\section{Applications of IFIB}
\label{app:solve_two_tasks}
In this section, we discuss how IFIB-C and IFIB-N solve the time-event prediction problem and event-time prediction problem in \cref{sec:application}.

\subsection{Applications with IFIB-C}
\label{sec:ifib-based_solution}
To answer the time-event prediction problem with IFIB-C, we first obtain \(P^*(t)\), the probability of an event happening between \(t_l\) and \(t\), as follows:
\begin{equation}
\begin{aligned}
\label{eqn:pred_t1}
    P^*(t) &= 1 - \sum_{m \in \mathrm{M}}{\int_{t}^{+\infty}{p^*(m, \tau)d\tau}} \\
    %m_p &= \argmax_{m}{p^*(m|t)} \\%= \argmax_{m}{\frac{ \frac{\partial \operatorname{IFIB}(m, t, \mathcal{H}_{t_l})}{\partial t} }{ \sum_{n \in M}{\frac{\partial \operatorname{IFIB}(m, t, \mathcal{H}_{t_l})}{\partial t}} }} \\
    %&= \argmax_{m}{ \frac{p^*(m, t)}{\sum_{n \in M}p^*(n, t)} } \\
    %&= \argmax_{m}{\frac{ \lambda^*(m, t)\exp(-\Lambda^*(t)) }{  \sum_{n \in M}{\lambda^*(n, t)\exp(-\Lambda^*(t))} }} \\
    %&= \argmax_{m}{ \frac{\lambda^*(m, t)}{\sum_{n \in M}{\lambda^*(n, t)}} }
\end{aligned}
% \label{eqn:pred_t}
%     P^*(t) &= 1 - \sum_{m \in M}{\operatorname{IFIB}(m, t, \mathcal{H}_{t_l})} \\
%     m_p &= \argmax_{m}{p^*(m|t)} = \argmax_{m}{\frac{ \frac{\partial \operatorname{IFIB}(m, t, \mathcal{H}_{t_l})}{\partial t} }{ \sum_{n \in M}{\frac{\partial \operatorname{IFIB}(m, t, \mathcal{H}_{t_l})}{\partial t}} }} \\
%     &= \argmax_{m}{ \frac{p^*(m, t)}{\sum_{n \in M}p^*(n, t)} } \\
%     &= \argmax_{m}{\frac{ \lambda^*(m, t)\exp(-\Lambda^*(t)) }{  \sum_{n \in M}{\lambda^*(n, t)\exp(-\Lambda^*(t))} }} \\
%     &= \argmax_{m}{ \frac{\lambda^*(m, t)}{\sum_{n \in M}{\lambda^*(n, t)}} }
% \end{aligned}
\end{equation}
For the time-event prediction problem, we follow the criterion in \cite{omi_fully_2019} by selecting the minimum \(t_p\) that allows \(P^*(t_p) \geqslant 0.5\). Because \(P^*(t)\) is a monotonical function, we could efficiently obtain \(t_p\) using the bisection method. Then, the mark of the next event at $t_p$ can be predicted. 
\begin{equation}
\begin{aligned}
\label{eqn:pred_t2}
    m_p &= \argmax_{m}{p^*(m|t_p)}
    = \argmax_{m}{ \frac{p^*(m, t_p)}{\sum_{n \in \mathrm{M}}p^*(n, t_p)}} 
\end{aligned}
\end{equation}

For the event-time prediction problem, we require the probability of an event marked by \(m\) happening between \(t_l\) and \(t\) denoted as \(P^*(t|m)\). It involves \(P^*(m)\), the integral of \(p^*(m, t)\) from \(t_l\) to positive infinity, and \(P^*(m, t)\), the integral of \(p^*(m, t)\) from \(t_l\) to \(t\) given \(m\). \cref{eqn:pred_time} shows the expression of \(P^*(t|m)\).
\begin{equation}
\begin{aligned}
%    m_p &= \argmax_{m}{p^*(m)} = \frac{\operatorname{IFIB}(m, t_l, \mathcal{H}_{t_l})}{\sum_{n \in M}{\operatorname{IFIB}(n, t_l, \mathcal{H}_{t_l})}} \\
\label{eqn:pred_time}
    P^*(t|m) = \frac{P^*(m, t)}{P^*(m)} = \frac{1}{\int_{t_l}^{+\infty}{p^*(m, \tau)d\tau}}\int_{t_l}^{t}{p^*(m, \tau)d\tau} \\
\int_{t_l}^{t}{p^*(m, \tau)d\tau} = \int_{t_l}^{+\infty}{p^*(m, \tau)d\tau} - \int_{t}^{+\infty}{p^*(m, \tau)d\tau}
\end{aligned}
\end{equation}
For each mark $m$, resembling how we obtain \(t\) from \cref{eqn:pred_t2}, the minimum $t_p$ such that $P^*(t_p|m) \geqslant 0.5$ is reported as the expected time that the next event will occur on the condition that the mark is $m$. Since \(P^*(t|m)\) is monotonic, it is also solvable by the bisection method. %It is useful to have \cref{eqn:pred_time}. For example, 
Meanwhile, we can predict which mark is more likely to be the next event.
\begin{equation}
\begin{aligned}
\label{eqn:pred_time1}
    m_p &= \argmax_{m}{p^*(m)} = \argmax_{m}{ \int_{t_l}^{+\infty}{p^*(m, \tau)}d\tau }
\end{aligned}
\end{equation}
Once $m_p$ is known, the time of the next event can be estimated using \cref{eqn:pred_time}. 

\subsection{Applications with IFIB-N}
IFIB-N solves the time-event prediction problem in a similar way as IFIB-C. It starts from obtaining \(P^*(t)\), the probability that an event with whatever mark could happen in the time interval $(t_l, t)$, as follows:
\begin{equation}
P^*(t) = 1 - \int_{t}^{+\infty}{\int_{\mathbf{r}\in \mathbf{M}}{p^*(\mathbf{r},\tau)d\tau d\mathbf{r}}}
\end{equation}
We then follow the criterion proposed by \cite{omi_fully_2019}, determining that an event happens once \(P^*(t_p) \geqslant 0.5\). \(P^*(t)\) is monotonic, so we could employ the bisect method to solve \(t_p\) efficiently. As for what mark this event would be, we probe the probability distribution \(p^*(\mathbf{m}, t_p)\) in the $n$-dimensional continuous space and output the mark $\mathbf{m}$ so that \(p^*(\mathbf{m}, t)\) hits the maximum. 
\begin{equation}
\mathbf{m}_p = \argmax_{\mathbf{m}}{p^*(\mathbf{m}, t_p)}
\label{eqn:find_m_p_1}
\end{equation}
Solving the event-time prediction problem with IFIB-N encounters an issue that the value of \(p(\mathbf{m}) = \int_{t_l}^{t}{p^*(\mathbf{m}, \tau)d\tau}\) at every \(\mathbf{m}\) is infinitesimal because there are infinite marks in the $n$-dimensional continuous space. Computers have difficulty handling infinitesimal. So, we replace \(p(\mathbf{m})\) with the probability in a small hypercube \(\mathcal{C}(\mathbf{m})\) in $\mathbf{M}$. This cube centers at \(\mathbf{m}\) with a very small edge. Now we can write the expression of \(P^*(t|\mathbf{m})\) as follows:
\begin{equation}
\label{eqn:cifib_p_t_m}
% \begin{aligned}
P^*(t|\mathbf{m}) = \frac{P^*(\mathbf{m}, t)}{P^*(\mathbf{m})} \\
= 1- \frac{\int_{t}^{+\infty}\int_{\mathbf{r}\in \mathcal{C}(\mathbf{m})}{p^*(\mathbf{r},\tau)}d\tau d\mathbf{r}}{\int_{t_l}^{+\infty}\int_{\mathbf{r}\in \mathcal{C}(\mathbf{m})}{p^*(\mathbf{r},\tau)}d\tau d\mathbf{r}} \\
 % %\int_{t}^{+\infty}\int_{r\in \mathcal{C}(\mathbf{m})}{p^*(\mathbf{r},\tau)}d\tau d\mathbf{r}}{1}%{\int_{t_l}^{+\infty}\int_{r\in \(\mathcal{C}(\mathbf{m})\)}p^*(\mathbf{r},\tau)d\tau d\mathbf{r}}
 % &= \frac{\int_{t_l}^{+\infty}{\int_{(m_1 - \Delta, \cdots, m_n - \Delta)}^{(m_1 + \Delta, \cdots, m_n + \Delta)}}{p^*(\mathbf{r}, t)d\mathbf{r}d\tau} - \int_{t}^{+\infty}{\int_{(m_1 - \Delta, \cdots, m_n - \Delta)}^{(m_1 + \Delta, \cdots, m_n + \Delta)}}{p^*(\mathbf{r}, t)d\mathbf{r}d\tau}}{\int_{t_l}^{+\infty}{\int_{(m_1 - \Delta, \cdots, m_n - \Delta)}^{(m_1 + \Delta, \cdots, m_n + \Delta)}}{p^*(\mathbf{r}, t)d\mathbf{r}d\tau}} \\
% &= 1 - \frac{\int_{t}^{+\infty}{p^*(\tau)d\tau \int_{m_1 - \Delta}^{m_1 + \Delta}{p^*(r_1|\tau)d r_1 \int_{m_2 - \Delta}^{m_2 + \Delta}{p^*(r_2|\tau, r_1)dr_2}\cdots \int_{m_n - \Delta}^{m_n + \Delta}{p^*(r_n|\tau, r_1, r_2, \cdots, r_{n - 1})dr_n}}}} {\int_{t_l}^{+\infty}{p^*(\tau)d\tau \int_{m_1 - \Delta}^{m_1 + \Delta}{p^*(r_1|\tau)d r_1 \int_{m_2 - \Delta}^{m_2 + \Delta}{p^*(r_2|\tau, r_1)dr_2}\cdots \int_{m_n - \Delta}^{m_n + \Delta}{p^*(r_n|\tau, r_1, r_2, \cdots, r_{n - 1})dr_n}}}} \\ 
% \end{aligned}
\end{equation}
For each probed mark \(\mathbf{m}\), we apply bisect method to find the minimum \(t_p\) so that \(P^*(t_p|\mathbf{m}) \geqslant 0.5 \). Meanwhile, we can effectively find the mark \(\mathbf{m}_p\) of the next event by, resembling \cref{eqn:find_m_p_1}, probing where \(p^*(\mathbf{m}, t_l)\) hits its maximum.
\begin{equation}
\mathbf{m}_p = \argmax_{\mathbf{m}}p^*(\mathbf{m}, t_l) = \argmax_{\mathbf{m}}{\int_{t_l}^{+\infty}{p^*(\mathbf{m}, \tau)d\tau}}
\end{equation}
After replacing \(\mathbf{m}\) in \cref{eqn:cifib_p_t_m} with \(\mathbf{m}_p\) and solving \(t_p\), we figure out the event-time prediction task by IFIB-N.

\begin{table*}[t]
\centering
\caption{Performance on synthetic datasets (higher Spearman, lower \(L^1\), and lower relative NLL loss are better.)}
\label{table:synthetic}
\vskip 0.15in
\begin{footnotesize}
\begin{tabular}{llccccc}
    \toprule
     & & Hawkes\_1 & Hawkes\_2 & Poisson & Self-correct & Stationary Renewal \\
    \midrule
\multirow{7}{*}{\rotatebox[origin=c]{90}{Spearman}} & IFIB-C(Ours) & \textbf{1.0000\tiny{\(\pm\)0.0000}} & \textbf{0.9999\tiny{\(\pm\)0.0000}} & \textbf{1.0000\tiny{\(\pm\)0.0000}} & \underline{0.9551\tiny{\(\pm\)0.0009}} & \textbf{0.9999\tiny{\(\pm\)0.0000}} \\
                          & FENN & 0.9946\tiny{\(\pm\)0.0004} & \underline{0.9964\tiny{\(\pm\)0.0002}} & 0.9736\tiny{\(\pm\)0.0006} & 0.9473\tiny{\(\pm\)0.0010} & 0.9998\tiny{\(\pm\)0.0000} \\
                          & FullyNN & 0.9952\tiny{\(\pm\)0.0004} & 0.9963\tiny{\(\pm\)0.0002} & 0.9722\tiny{\(\pm\)0.0018} & 0.9477\tiny{\(\pm\)0.0001} & 0.9998\tiny{\(\pm\)0.0000} \\
                          & RMTPP & 0.9839\tiny{\(\pm\)0.0000} & 0.8473\tiny{\(\pm\)0.0001} & \textbf{1.0000\tiny{\(\pm\)0.0000}} & \textbf{0.9557\tiny{\(\pm\)0.0001}} & 0.0204\tiny{\(\pm\)0.0000} \\
                          & LogNormMix & 0.9816\tiny{\(\pm\)0.0000} & 0.9772\tiny{\(\pm\)0.0000} & 0.9685\tiny{\(\pm\)0.0002} & 0.9462\tiny{\(\pm\)0.0000} & \textbf{0.9999\tiny{\(\pm\)0.0000}} \\
                          & SAHP & \underline{0.9959\tiny{\(\pm\)0.0047}} & 0.9862\tiny{\(\pm\)0.0000} & 0.9615\tiny{\(\pm\)0.0025} & 0.9492\tiny{\(\pm\)0.0014} & 0.9990\tiny{\(\pm\)0.0007} \\
                          & THP & 0.9266\tiny{\(\pm\)0.0026} & 0.7366\tiny{\(\pm\)0.0005} & \textbf{1.0000\tiny{\(\pm\)0.0000}} & 0.6969\tiny{\(\pm\)0.0017} & 0.0413\tiny{\(\pm\)0.0024} \\
    \midrule
\multirow{7}{*}{\rotatebox[origin=c]{90}{\(L^1\)}}  & IFIB-C(Ours) & \textbf{0.1480\tiny{\(\pm\)0.0085}} & \textbf{0.3105\tiny{\(\pm\)0.0432}} & \textbf{0.0133\tiny{\(\pm\)0.0091}} & \underline{0.7907\tiny{\(\pm\)0.0711}} & \underline{ 0.0654\tiny{\(\pm\)0.0018}} \\
                          & FENN & 0.6248\tiny{\(\pm\)0.0052} & \underline{3.0398\tiny{\(\pm\)0.0693}} & 0.2919\tiny{\(\pm\)0.0051} & 1.2139\tiny{\(\pm\)0.1652} & 0.0703\tiny{\(\pm\)0.0058} \\
                          & FullyNN & \underline{0.6235\tiny{\(\pm\)0.0227}} & 3.1048\tiny{\(\pm\)0.0763} & 0.2973\tiny{\(\pm\)0.0098} & 1.1889\tiny{\(\pm\)0.0244} & 0.0710\tiny{\(\pm\)0.0099} \\
                          & RMTPP & 5.1690\tiny{\(\pm\)0.0028} & 21.144\tiny{\(\pm\)0.0115} & \underline{0.0469\tiny{\(\pm\)0.0111}} & \textbf{0.7193\tiny{\(\pm\)0.0329}} & 9.7272\tiny{\(\pm\)0.0031} \\
                          & LogNormMix & 0.8767\tiny{\(\pm\)0.0038} & 4.2257\tiny{\(\pm\)0.0338} & 0.3861\tiny{\(\pm\)0.0036} & 1.1145\tiny{\(\pm\)0.0301} & \textbf{0.0405\tiny{\(\pm\)0.0056}} \\
                          & SAHP & 1.0245\tiny{\(\pm\)0.2967} & 4.7867\tiny{\(\pm\)0.2735} & 0.6893\tiny{\(\pm\)0.0238} & 1.3363\tiny{\(\pm\)0.0196} & 0.4872\tiny{\(\pm\)0.1833} \\
                          & THP & 12.003\tiny{\(\pm\)0.2069} & 25.500\tiny{\(\pm\)0.3642} & 0.0203\tiny{\(\pm\)0.0067} & 10.656\tiny{\(\pm\)0.0965} & 9.9230\tiny{\(\pm\)0.0451} \\
    \midrule
\multirow{7}{*}{\rotatebox[origin=c]{90}{Relative NLL}} & IFIB-C(Ours) & \textbf{0.0000\tiny{\(\pm\)0.0000}} & \textbf{0.0001\tiny{\(\pm\)0.0000}} & \textbf{0.0000\tiny{\(\pm\)0.0000}} & \underline{0.0007\tiny{\(\pm\)0.0003}} & \textbf{0.0000\tiny{\(\pm\)0.0000}} \\
                          & FENN &0.0003\tiny{\(\pm\)0.0000} & 0.0009\tiny{\(\pm\)0.0001} & 0.0002\tiny{\(\pm\)0.0000} & 0.0016\tiny{\(\pm\)0.0006} & \textbf{0.0000\tiny{\(\pm\)0.0000}} \\
                          & FullyNN &0.0003\tiny{\(\pm\)0.0000} & 0.0008\tiny{\(\pm\)0.0001} &0.0002\tiny{\(\pm\)0.0000} & 0.0015\tiny{\(\pm\)0.0001} & \textbf{0.0000\tiny{\(\pm\)0.0000}} \\
                          & RMTPP &0.0586\tiny{\(\pm\)0.0000} & 0.3653\tiny{\(\pm\)0.0000} & \textbf{0.0000\tiny{\(\pm\)0.0000}}  & \textbf{0.0003\tiny{\(\pm\)0.0000}} & 0.0615\tiny{\(\pm\)0.0000} \\
                          & LogNormMix & \underline{0.0001\tiny{\(\pm\)0.0000}} & \underline{0.0002\tiny{\(\pm\)0.0000}} & \textbf{0.0000\tiny{\(\pm\)0.0000}} & 0.0010\tiny{\(\pm\)0.0000} & \textbf{0.0000\tiny{\(\pm\)0.0000}} \\
                          & SAHP & 0.0086\tiny{\(\pm\)0.0017} & 0.0312\tiny{\(\pm\)0.0193} & 0.0092\tiny{\(\pm\)0.0002} & 0.0072\tiny{\(\pm\)0.0009} & 0.0034\tiny{\(\pm\)0.0010} \\
                          & THP & 0.2137\tiny{\(\pm\)0.0001}& 0.6663\tiny{\(\pm\)0.0029} & \textbf{0.0000\tiny{\(\pm\)0.0000}} & 0.1262\tiny{\(\pm\)0.0004} & 0.0771\tiny{\(\pm\)0.0000} \\
    \bottomrule
\end{tabular}
\end{footnotesize}
\vskip -0.1in
\end{table*}

\section{Experiment results on synthetic datasets}
\subsection{Results on IFIB-C}
\label{app:synthetic_cate_results}
Synthetic datasets can help us verify whether a marked TPP model works correctly. % by measuring the gap between \(p^*(m, t)\) and \(\hat{p}^*(m, t)\). %In this section, we train different marked TPP models on the synthetic datasets and report the Spearman coefficient, \(L^1\) distance, and relative NLL loss between \(p^*(m, t)\) and \(\hat{p}^*(m, t)\).\par
%As aforementioned, 
To this end, we compare the real \(p^*(m, t)\) of each synthetic dataset with \(\hat{p}^*(m, t)\) modeled using IFIB-C and baselines. The selected metrics are Spearman's coefficient, \(L^1\) distance, and relative NLL loss. The results are reported in \cref{table:synthetic}. The numbers in bold or underlined indicate the best or the second-best value.

% Additionally, we use the mark "R,  referring to RNN and "T", meaning Transformers, to emphasise the disparity in the history encoder between SAHP, THP and other marked TPP models. 
IFIB-C, FullyNN, FENN, SAHP, and LogNormMix can consistently return correct \(p^*(m, t)\) in most situations as their Spearman's coefficient is close to 1, and their \(L^1\) distance is also acceptable. On the other hand, RMTPP and THP struggle to learn the correct distribution on some synthetic datasets. The limitation of the presumed fixed-form intensity function may cause this. 

\begin{figure*}[ht]
    \centering
    \begin{subfigure}{0.24\textwidth}
        \includegraphics[width=\textwidth]{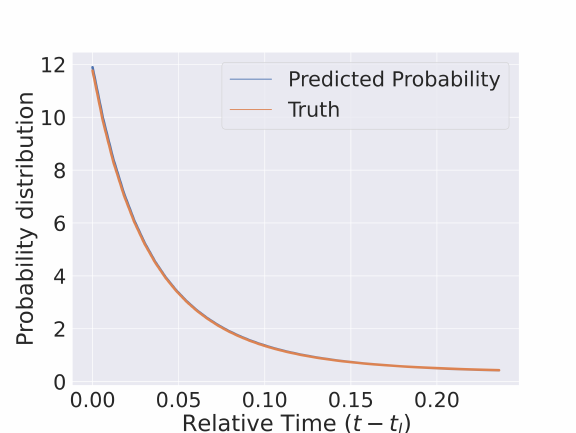}
        \label{fig:IFIB_probe}
        \caption{IFIB}
    \end{subfigure}
    \begin{subfigure}{0.24\textwidth}
        \includegraphics[width=\textwidth]{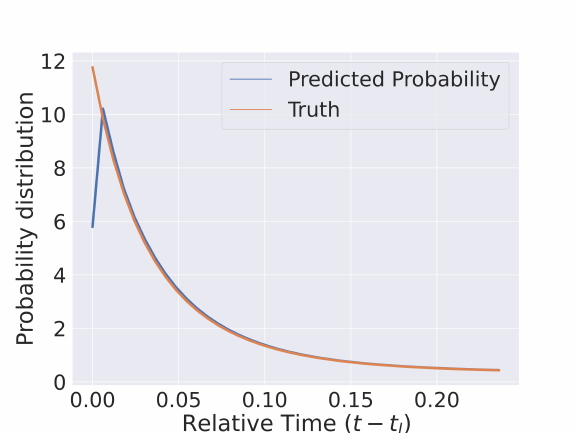}
        \label{fig:fenn_probe}
        \caption{FENN}
    \end{subfigure}
    \begin{subfigure}{0.24\textwidth}
        \includegraphics[width=\textwidth]{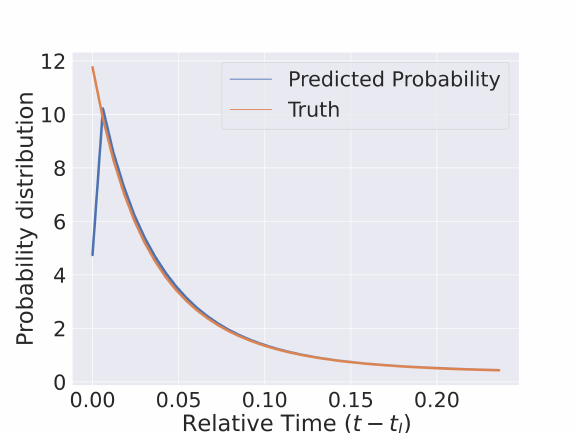}
        \label{fig:fullynn_probe}
        \caption{FullyNN}
    \end{subfigure} 
    \begin{subfigure}{0.24\textwidth}
        \includegraphics[width=\textwidth]{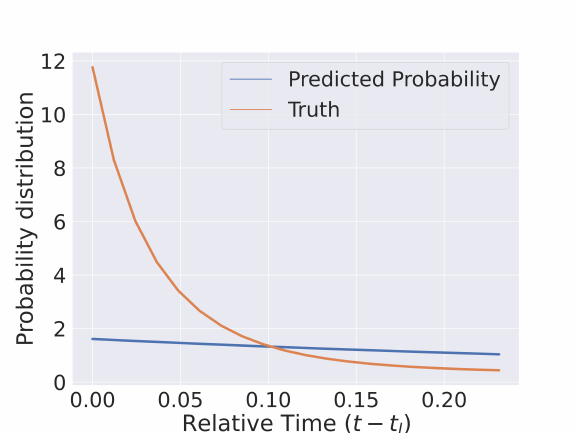}
        \label{fig:rmtpp_probe}
        \caption{RMTPP}
    \end{subfigure}
    \begin{subfigure}{0.24\textwidth}
        \includegraphics[width=\textwidth]{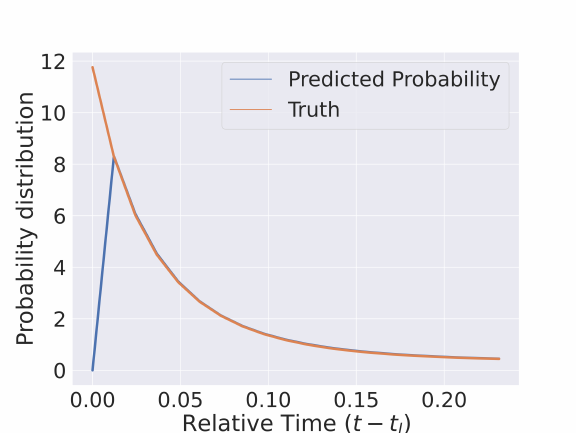}
        \label{fig:ifl_probe}
        \caption{LogNormMix}
    \end{subfigure}
    \begin{subfigure}{0.24\textwidth}
        \includegraphics[width=\textwidth]{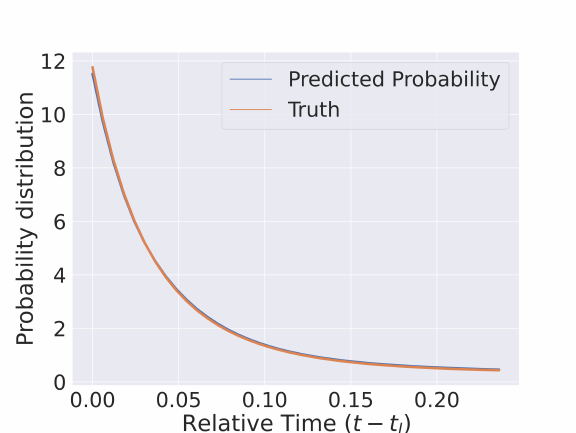}
        \label{fig:sahp_probe}
        \caption{SAHP}
    \end{subfigure}
    \begin{subfigure}{0.24\textwidth}
        \includegraphics[width=\textwidth]{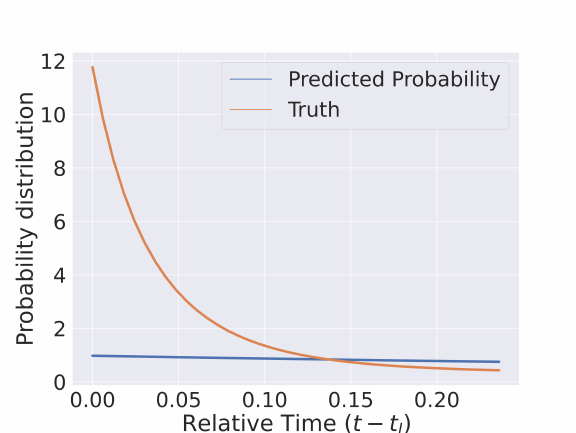}
        \label{fig:thp_probe}
        \caption{THP}
    \end{subfigure}
    \caption{Curves of \(\hat{p}^*(m, t)\) and \(p^*(m, t)\) around \(t_l\) on Hawkes\_2. We notice that FullyNN, FENN, and LogNormMix often start from an unusually low value at \(t_l\) before returning to the correct values, causing a high \(L^1\) distance and low Spearman's coefficient. The relative NLL loss, on the other hand, cannot reflect this phenomenon in its value.}
    \label{fig:probe}
\end{figure*}

The results demonstrate that IFIB-C always learns marked TPP models with high fidelity, which consistently have the best \(L^1\) distance, negative NLL loss, and Spearman's coefficient in almost all cases. Only RMTPP and Lognormmix defeat IFIB-C on the self-correct and stationary renewal dataset. FENN and FullyNN also perform strongly from the perspective of relative NLL loss, but the values in \(L^1\) distance reveal that FENN and FullyNN cannot perfectly estimate \(p^*(m, t)\) at every \(t\). %This is affirmed after scrutinising the estimated probability distribution \(\hat{p}^*(m, t)\) as 
Specifically, \(\hat{p}^*(m, t)\) around \(t_l\) might deviate from \(p^*(m, t)\). In contrast, IFIB-C successfully avoids this issue. This can be observed clearly when we compare \(\hat{p}^*(m, t)\) and \(p^*(m, t)\) at every time in a short time period after $t_l$, as shown in \cref{fig:probe} where $0$ in x-axis is $t_l$. More experiments (not reported) prove that such deviations exist in all synthetic datasets.
%Figures are available in \cref{app:dist} and \cref{app:plot_synthetic}.\par
 
\subsection{Results on IFIB-N}
\label{app:synthetic_num_results}
\cref{tab:cifib_synthetic_performance} reports IFIB-N's performance on five synthetic datasets. The metric is Spearman coefficient, \(L^1\) distance, and relative NLL loss. As illustrated, IFIB-N performs as well as IFIB-C on synthetic datasets in terms of Spearman coefficient and \(L^1\) distance. It proves that IFIB-N can learn the correct distribution with high fidelity. The relative NLL loss is not reported since it is unsuitable when marks are continuous, 

%But IFIB-N has a much higher relative NLL loss compared with IFIB-C. The explanation is the relative NLL loss is calculation process on discrete models does not work well for TPP with the continuous marks.

\begin{table}[ht]
    \centering
    \caption{Performance of IFIB-N on synthetic datasets (higher Spearman and lower \(L^1\) are better.)}
    \vskip 0.15in
    \begin{tabular}{lccccc}
        \toprule
                 & Hawkes\_1 & Hawkes\_2 & Poisson & Self-correct & Stationary Renewal \\
        \midrule
        Spearman & 1.0000\tiny{\(\pm\)0.0000} & 1.0000\tiny{\(\pm\)0.0000} & 1.0000\tiny{\(\pm\)0.0000} & 0.9647\tiny{\(\pm\)0.0001} & 0.9998\tiny{\(\pm\)0.0001} \\
        \(L^1\) & 0.1186\tiny{\(\pm\)0.0040} & 0.2042\tiny{\(\pm\)0.0135} & 0.0415\tiny{\(\pm\)0.0108} & 0.5451\tiny{\(\pm\)0.0262} & 0.0960\tiny{\(\pm\)0.0225} \\
        %Relative NLL & 1.4310\tiny{\(\pm\)0.0017} & 1.4326\tiny{\(\pm\)0.0036} & 1.3833\tiny{\(\pm\)0.0006} & 1.4449\tiny{\(\pm\)0.0084} & 1.4313\tiny{\(\pm\)0.0022} \\
        \bottomrule
    \end{tabular}
    \label{tab:cifib_synthetic_performance}
    \vskip -0.1in
\end{table}

\section{Experiment Settings}
\label{app:hyperparameters}
%This section introduces data preprocessing and the parameters of each solution, including model structure (MS), batch size (BS), and learning rate (LR). The two values of "Steps" refers to the number of warm-up steps and total training steps. While others are the same in general, different solutions have different settings on MS.  

\subsection{Datasets}
\label{app:datasets}
%This section provides detailed information about synthetic and real-world datasets used in this paper.
\subsubsection{Real-world Datasets}
\label{app:realworld_generation}
The following four datasets contain discrete marks. We use these datasets to evaluate IFIB-C.
\begin{itemize}
    \item \textit{Bookorder dataset} (BO) \cite{du_recurrent_2016} logs the frequent stock transactions from NYSE. Each event (i.e., transaction) belongs to one of the two event types\footnote{In this study, event types and event marks are equivalent}: buy or sell. The number of events is 400K, and the average sequence length is 3,319 for the training and evaluation set and 829 for the test set. 
    \item \textit{Retweet dataset} \cite{zhao_seismic_2015} records when users retweet a particular message on Twitter. This dataset distinguishes all users into three different types: (1) normal user, whose followers count is lower than the median, (2) influence user, whose followers count is higher than the median but lower than the $95$th percentile, (3) famous user, whose followers count is higher than the $95$th percentile. About 2 million retweets are recorded, and the average sequence length is 108.
    \item \textit{StackOverflow dataset} (SO) \cite{Leskovec2014SNAPD} was collected from Stackoverflow\footnote{https://stackoverflow.com/}, a popular question-answering website about various topics. Users providing decent answers will receive different badges as rewards. This dataset collects the timestamps when people obtain 22 badges from the website, and the average sequence length is 72.
    \item \textit{MOOC} \cite{shchur_intensity-free_2020} records users' interactions on an online course website. These actions include watching online courses, sitting for a quiz, or interacting with other students or teachers. The average length of sequences in this dataset is 57, and as many as 97 marks are available. %Such a relatively large amount of marks is a real challenge to all marked TPP modeling, especially one learning \(p^*(m, t)\) or \(\lambda^*(m, t)\).
\end{itemize}

The following three datasets have continuous marks. We use these datasets to evaluate the performance of IFIB-N. All three datasets come from \cite{chen_neural_2020}. 
\begin{itemize}
    \item \textit{Citibike} records bike usage behaviors in New York City. There are 6,240 record sequences in the training set, 500 in the validation set, and 500 in the test set. The length of these sequences varies from 9 to 231, and the average length is 113.
    \item \textit{COVID-19} records the daily COVID-19 cases in New Jersey State published by the New York Times\footnote{https://github.com/nytimes/covid-19-data}. The training set has 2,050 records, and both the validation and test datasets have 100 sequences. The number of sequences ranges from 3 to 323.
    \item \textit{Earthquake} contains the time and position of earthquakes and their aftershocks with a magnitude of at least 2.5 in Japan from 1990 to 2020, detected and recorded by the U.S. Geological Survey\footnote{https://earthquake.usgs.gov/earthquakes/search/}. The number of sequences in the training, validation, and test set is 1,300, 100, and 100, respectively, and the length of sequences ranges from 19 to 545.
\end{itemize}

\subsubsection{Synthetic Datasets}
\label{app:synthetic_generation}

\textit{Hawkes process dataset} was generated utilising two different hawkes processes: Hawkes\_1: \(\lambda^{*}(t) = \mu_0 + \sum_{t_i < t}{a\exp(-b(t - t_i))}\) where \(\mu = 0.2\), \(a = 0.8\), and \(b = 1.0\), and 2. Hawkes\_2: \(\lambda^{*}(t) = \mu_0 + \sum_{t_i < t}{a_1\exp(-b_1(t - t_i)) + a_2\exp(-b_2(t - t_i))}\) where \(\mu = 0.2\), \(a_1 = a_2 = 0.4\), \(b_1 = 1.0 \), and \(b_2 = 20\).

\textit{Homogeneous Poisson process dataset} was generated using the Homogeneous Poisson process where the conditional intensity function \(\lambda^*(t)\) is constant over the entire timeline. This paper assumes \(\lambda^*(t) = 1\) for the discrete-marked dataset and \(\lambda^*(t) = 0.75\) for the continuous-marked one.

\textit{Self-correct process dataset} was generated using the temporal point process whose intensity significantly drops when an event happens. The definition of the conditional intensity function is \(\lambda^*(t) = \exp(\mu * (t - t_i) - \alpha * N)\) where \(N\) is the number of occurred events, and \(\mu\) and \(\alpha\) are fixed parameters. In our experiments, we set \(\alpha = \mu = 1\) for the discrete-marked dataset and \(\alpha = 1, \mu = 1.5\) for the continuous-marked one.

\textit{Stationary renewal process dataset} was generated using stationary renewal process, which directly defines the probability distribution over time \(p^*(t)\) as a log-normal distribution as shown in \cref{eqn:log_norm}.
\begin{equation}
\label{eqn:log_norm}
p^*(t|\sigma) = \frac{1}{\sigma t\sqrt{2\pi}}\exp(-\frac{\log^2(t)}{2\sigma^2})
\end{equation}
where \(\sigma\) is the standard deviation. Here, we set \(\sigma = 1\). With \cref{eqn:log_norm} and TPP's definition, one could solve the corresponding intensity function by Wolframalpha\footnote{https://www.wolframalpha.com}:
\begin{equation}
\label{eqn:intensity}
    \lambda^*(t) = \frac{-0.797885*\exp(-0.5*\log^2(t))}{-t + t * \operatorname{erf}(0.707107 * \log(t))}
\end{equation}
where \(\operatorname{erf}(x) = \frac{2}{\sqrt{\pi}}\int_{0}^{x}{\exp(-t^2)dt}\). %This process is significantly more challenging to estimate than other synthetic processes mentioned beforehand, so one might train MFullyNN and FullyNN-ETE with different random seeds to obtain a working model.

Similar to the real-world datasets setup, these five synthetic distributions will cooperate with two synthetic marking methods. One generates discrete marks sampled from a uniform distribution with \(N = 5\) options, and the other creates continuous marks from a spinning wheel distribution with \(N = 7\) leaves, as shown in \cite{chen_neural_2020}.

\subsection{Metric}
\label{app:metric}
\subsubsection{Metrics for Synthetic Datasets}
%However, MAE is event-agnostic and cannot be applied directly to evaluate event-oriented marked TPP. For each marker whose probability of being the next event is non-zero, the event-oriented marked TPP model predicts the time of the next event on the condition that the next event is this marker. That is, there is a time prediction for each of such markers. It does not make sense to compare the time prediction for one maker against the real-time of the next event, which is another marker. Therefore, we adapt MAE to MAE-E (MAE by event). Suppose the next event is marker $m$. MAE-E is the absolute value of the predicted time of the next event for mark $m$ and the real-time of the next event, i.e., $\operatorname{MAE-E} = |\hat{t}_m - t|$.

For synthetic datasets, the real $p^*(m, t)$ is known. We can compare the generated $\hat{p}^*(m, t)$ against the real one. A smaller discrepancy between $\hat{p}^*(m, t)$ and $p^*(m, t)$ indicates the marked TPP has been better modelled. %It gives us the confidence to use the method to model the marked TPP on real-world datasets. 
Most papers report the relative NLL loss, that is, the average of the absolute difference between \(-\log\hat{p}^*(m, t)\) and \(-\log{p}^*(m, t)\), or \(-\log\hat{p}^*(t)\) and \(-\log{p}^*(t)\) if markers are unavailable in the synthetic datasets, at event points as the discrepancy \cite{omi_fully_2019, shchur_intensity-free_2020}. The lower relative NLL loss indicates a better performance. However, such a metric only evaluates %ensures satisfactory 
performance at discrete event points, which cannot gauge the overall discrepancy between $\hat{p}^*(m, t)$ and $p^*(m, t)$. To fill the gap, % truly measure the discrepancy between $\hat{p}^*(m, t)$ and $p^*(m, t)$, we need new metrics that minds the differences between $\hat{p}^*(m, t)$ and $p^*(m, t)$ involving all available timestamps. 
this paper selects Spearman's Coefficient \(\rho\) and \(L^1\) distance to measure the discrepancy between $\hat{p}^*(m, t)$ and $p^*(m, t)$ over time, while we also report the relative NLL loss for reference.

\textit{Spearman's Coefficient} $\rho(X, Y)$ measures the relationship between two arbitrary value sequences, \(X\) and \(Y\), as defined by \cref{eqn:spearman}. %Let $X$ and $Y$ are the values of $f(x)$ and $g(x)$, respectively, for all $x$ in \([a, b]\).
 If \(X\) and \(Y\) are more correlated, $\rho(X, Y)$ is higher; lower otherwise. Compared with the Pearson coefficient which is suitable if the relationship between \(X\) and \(Y\) is linear, Spearman's coefficient could better deal with non-linear relationships. Because most probability distributions of TPP are non-linear, we select Spearman's coefficient.
\begin{equation}
\label{eqn:spearman}
    \rho(X, Y) = \frac{\operatorname{Cov}(\operatorname{Rank}(X), \operatorname{Rank}(Y))}{\sigma_{X}\sigma_{Y}} \in [-1, 1]
\end{equation}
where \(\sigma_{X}\) and \(\sigma_{Y}\) are the standard deviations of the values in sequence \(X = \{x_1, x_2, \cdots, x_n\}\) and \(Y = \{y_1, y_2, \cdots, y_n\}\), respectively. We expect \(\rho\) between $\hat{p}^*(m, t)$ and $p^*(m, t)$ is close to 1.

\textit{\(L^1\) distance} measures how different two arbitrary functions are in interval \([a, b]\).
\begin{equation}
    L^1(f, g) = \int_{a}^{b}{|f(x) - g(x)|dx} \geqslant 0
\end{equation}
The smaller the \(L^1\) distance is, the more similar $f(x)$ and $g(x)$ are. When \(L^1(f, g) = 0\), \(f(x)\) almost equals to \(g(x)\) in interval \([a, b]\) for any \(f(x)\) and \(g(x)\), or \(f(x) = g(x)\) at every \(x \in [a, b]\) if both \(f(x)\) and \(g(x)\) are continuous.

%\subsection{Definition of other metrics}
%In this section, we elaborate the remaining  metrics, macro-F1, for measuring the event prediction performance on all datasets. 

\subsubsection{Metrics for Real-world Datasets}
\textit{MAE} (Mean Absolute Error) is the absolute value mean of prediction and the ground truth. MAE has been used to measure how good the model could be for predicting when the next event will happen without concerning the event mark. We notice that MAE can be remarkably affected by a few poor predictions. To measure prediction performance more reliably, we sort the prediction errors for all observations and report the $25$th percentile (Q1), $50$th percentile (Median), and $75$th percentile (Q3), denoted as MAE$@25\%$, MAE$@50\%$, MAE$@75\%$, respectively. Comparing the $x$th percentile of two methods, the one with the lower value has the better performance on the $x\%$ best predictions.    

\textit{MAE-E} (MAE by Event) %is the absolute va Since this paper concerns both time and the mark of the next event, we need to adapt MAE to MAE-E (MAE by event).
is a variant of MAE with consideration of mark information. Supposing the real mark of the next event is $m$. MAE-E is the absolute value mean of the next event time predicted for mark $m$ and the real-time of the next event. For the same reason, we sort the prediction errors for all observations and report the $25$th, $50$th, and $75$th percentile, denoted as MAE-E$@25\%$, MAE-E$@50\%$, MAE-E$@75\%$, respectively. The lower value indicates better performance. 

%\begin{equation}
%     \operatorname{MAE-E}(\hat{t_m}, t_m) = |\hat{t}_m - t_m|
% \end{equation}

\textit{Macro-F1} is derived from the F1 value. %F1 value has been widely used in almost all binary classification tasks because, compared with accuracy that might be fooled by false positives, F1 value takes accuracy and recall rate in its mind, where the model should correctly mark out positive samples for a better accuracy and negative samples for a better recall rate. 
The definition of F1 value is:
\begin{equation}
    \operatorname{F1} = \frac{2\times \operatorname{Accuracy} \times \operatorname{Recall}}{\operatorname{Accuracy} + \operatorname{Recall}}
\end{equation}
F1 value is only for binary classification, but researchers realize that a multi-class classification can be evaluated by decomposing the original classification task into multiple binary classification tasks. In this study, each mark $m$ is a class, and the binary classification predicts whether the next event is $m$. Let the F1 value of mark $m$ be $\operatorname{F1}_m$. The macro-F1 is defined in \cref{equ:marcof1}. The macro-F1 ranges from 0 to 1, where 0 is the worst possible score, and 1 is a perfect score indicating that the model predicts each observation correctly.
\begin{equation}\label{equ:marcof1}
    \operatorname{macro-F1} = \frac{1}{|M|}\sum_{m\in M}{\operatorname{F1}_m}
\end{equation}

\textit{DV}(Distance between the predicted vector and the ground truth) is the Euclidean distance between the predicted vector \(\hat{\mathbf{m}} = (\hat{d_1}, \hat{d_2}, \cdots, \hat{d_n})\) and the recorded ground truth \(\mathbf{m} = (d_1, d_2, \cdots, d_n)\), as shown in \cref{eqn:DV_def}.
\begin{equation}
    DV(\hat{\mathbf{m}}, \mathbf{m}) = \sqrt{\sum_{i = 1}^{n}{(\hat{d_i} - d_i)^2}}
    \label{eqn:DV_def}
\end{equation}
In this paper, we report the 25th percentile(Q1, denoted as DV\(@25\%\)), 50th percentile(Q2, as known as the median number, denoted as DV\(@50\%\)), and 75th percentile(Q3, denoted as DV\(@75\%\)) for more reliable prediction performance measurement. Like MAE and MAE-E, lower DV\(@x\%\) indicates better performance.

\subsection{Baseline Models}
\label{app:baseline}
This section provides further information about the baseline models used in the experiments for evaluating IFIB-C. 

\textbf{Recurrent Marked Temporal Point Process (RMTPP)}\cite{du_recurrent_2016} uses an RNN encoder to represent history as a hidden state \(\mathbf{h}\) based on which the intensity function and the density function are modelled to predict the time of next event. The intensity function is formulated as an exponential function. The RMTPP sets up a dedicated mark generation module to predict the mark of the next event so that it can solve the time-event prediction problem, but not the event-time prediction problem.% and mark evolution.
% while this paper patched RMTPP by duplicating the predefined intensity function, one for each marker, to enable native marked TPP learning. This patched RMTPP is only used in experiments involving real-world datasets.

\textbf{Fully Neural Network (FullyNN)} \cite{omi_fully_2019} uses a neural network to estimate the integral of $\lambda^*(t)$ for the history embedding \(\mathbf{h}\) and inter-event time $t$. Then the density function is formulated to predict the time of the next event. FullyNN is designed for TPP without the information of event marks. To work with marked TPP, FullyNN can be simply extended but it has a performance issue. Details are available in \cref{appendix:FullyNN}.

\textbf{Fully Event Neural Network (FENN)} is an extension of FullyNN by changing the shared \(\mathbf{v}\) in FullyNN to a set of event-specific vectors. FENN successfully overcomes the computation graph overlap issue yet still inherits FullyNN's drawback of failing to respect the mathematical restrictions. We believe sometimes such failure might be responsible for the inferior performance of FENN. Detailed information about FENN is available in \cref{apx:ETE}.

\textbf{ Transformer Hawkes Process (THP)} \cite{zuo_transformer_2020} follows the idea of RMTPP but employs the Transformer-based encoder for more powerful history embeddings and replaces the exponential function with a softplus-based intensity function.

\textbf{LogNormMix} \cite{shchur_intensity-free_2020} directly models the density function \(p^*(t)\) with a mixture of log-normal distributions. The density function is conditioned on history embedding \(\mathbf{h}\) provided by an RNN model. Akin to RMTPP, LogNormMix employs a dedicated module for \(p^*(m)\). Therefore, it only can solve the time-event prediction problem. More details of LogNormMix is in \cref{app:lognormmix}

% \textbf{Continuous-time LSTM (CTLSTM)}\cite{mei_neural_2017} 
% generalizes the classical Hawkes process by parameterizing its intensity function with recurrent neural networks. CLSTM is an interpolated version of the standard LSTM, allowing us to generate outputs in a continuous-time domain.

\textbf{Self-Attentive Hawkes Process (SAHP)} \cite{zhang_self-attentive_2020} is based on the same intuition as Continuous-time LSTM (CTLSTM) \cite{mei_neural_2017}, which generalizes the classical Hawkes process by parameterizing its intensity function with recurrent neural networks. CTLSTM is an interpolated version of the standard LSTM, allowing us to generate outputs in a continuous-time domain. SAHP further improves performance by replacing LSTM with Transformers. Because the only difference between SAHP and CTLSTM is the history encoder, and SAHP has reported achieving better performance than CTLSTM, we only evaluate SAHP in this paper.

% \textbf{Fully Event-aware Neural Network learned} is an intuitive marked TPP adaptation of FullyNN after realising the FullyNN cannot tackle marked TPP problems because of the computation graph overlap. Detailed information about this model is available in \cref{apx:ETE}.

\subsection{Data Preprocessing}
We prepare synthetic and real-world datasets with two preprocessing methods, i.e., normalization and inception reset. Normalization scales the times of events in the event sequences by their mean \(\bar{t}\) and standard deviation \(\sigma\). Inception reset ensures %shifts the first dummy event on the timeline so that 
the first real event always happens at a predefined time \(t_{\text{offset}}\). The former is useful when the time is relatively large, such as in the Retweet dataset. The latter is crucial when the original event sequence always starts from an unexpectedly large time, such as in the stackoverflow dataset. \cref{tab:datasets} shows how the two preprocessing methods are applied on various datasets\footnote{"Synthetic" refers to the five synthetic datasets in \cref{app:datasets}.}.

\begin{table}[!ht]
    \centering
    \caption{Data preprocessing.}
    \vskip 0.15in
    \begin{tabular}{ccc}
    \toprule
         Dataset name & Normalization & Resetted inception \\
    \midrule
         Retweet & \checkmark & \ding{55} \\
         StackOverflow & \checkmark & \checkmark(\(t_{\text{offset}} = 0.8\)) \\
         MOOC & \checkmark & \ding{55} \\
         Bookorder & \checkmark & \ding{55} \\
         COVID-19 & \ding{55} & \ding{55} \\
         Citibike & \ding{55} & \ding{55} \\
         Earthquakes & \ding{55} & \ding{55} \\
    \midrule
         Synthetic(with continuous and discrete marks) & \ding{55} & \ding{55} \\
    \bottomrule
    \end{tabular}
    \vskip -0.1in
    \label{tab:datasets}
\end{table}

\subsection{Model Training}
This section lists the hyperparameter settings of all TPP and marked TPP models used in this paper. The two values of "Steps" refers to the number of warm-up steps and total training steps. "BS" refers to batch size, and "LR" refers to the learning rate. Unless otherwise specified, we repeatedly train a model \(N = 3\) times on an NVIDIA A100-PCIE GPU and an NVIDIA P40 GPU(IFIB-N specific) with different random seeds and report the mean and standard variance of every result to aid any reproducibility concerns. We will release all model checkpoints used upon acceptance.

\subsubsection{IFIB-C Configurations}
\cref{tab:IFIB_hyp} lists the IFIB-C's hyperparameter settings. The three values of "MS" (model structure) refer to the number of dimensions for history embedding \(\mathbf{h}\), the number of dimensions for $\mathbf{f}(m, t)$, and the number of non-negative fully-connected layers in the IEM module, respectively. 

\begin{table}[!ht]
    \centering
    \caption{Hyperparameter settings for IFIB-C (\(N = 10\) for training IFIB-C on the Poisson dataset and \(N = 3\) on other datasets).}
    \label{tab:IFIB_hyp}
    \vskip 0.15in
    \begin{footnotesize}
    \begin{tabular}{ccccc}
    \toprule
      Datasets & Steps & MS & BS & LR \\
    \midrule
      Retweet & [80,000, 400,000] & [32, 64, 3] & 32 & 0.002 \\
      MOOC & [80,000, 400,000] & [32, 32, 2] & 32 & 0.002 \\
      SO & [40,000, 200,000] & [32, 32, 2] & 32 & 0.002 \\
      BO & [4,000, 20,000] & [32, 32, 2] & 8 & 0.002 \\
      Synthetic & [1,000, 10,000] & [32, 64, 3] & 128 & 0.002 \\
    \bottomrule
    \end{tabular}
    \vskip -0.1in
    \end{footnotesize}
\end{table}

% As for the scaling factors \(\alpha\) and \(\beta\) in \cref{eqn:IFIB_layer}, we first pretrain an IFIB-C model with trainable \(\alpha\) and \(\beta\) following the hyperparameter settings in \cref{tab:IFIB_hyp}. We find the checkpoint achieving the highest macro-F1. After pretraining, we extract the new scaling factors from the checkpoint and retrain the IFIB-C with the new scaling factors which are fixed. We have only used this trick for experiments on real-world datasets.

\subsubsection{FullyNN and FENN Configurations}
\cref{tab:FullyNN_ete_hyp} shows hyperparameter settings for FullyNN and FENN. The three numbers in column "MS" share the same meaning as in IFIB-C.

\begin{table}[!ht]
    \centering
    \caption{Hyperparameter settings for FullyNN and FENN.}
    \label{tab:FullyNN_ete_hyp}
    \vskip 0.15in
    \begin{footnotesize}
    \begin{tabular}{ccccc}
    \toprule
      Datasets & Steps & MS & BS & LR \\
    \midrule
      Retweet & [20,000, 100,000] & [32, 16, 4] & 32 & 0.002 \\
      MOOC & [80,000, 400,000] & [32, 32, 2] & 32 & 0.002 \\
      SO & [10,000, 50,000] & [32, 32, 2] & 32 & 0.002 \\
      BO & [4,000, 20,000] & [32, 32, 2] & 8 & 0.002 \\
      Synthetic & [1,000, 10,000] & [32, 64, 3] & 128 & 0.002 \\
    \bottomrule
    \end{tabular}
    \vskip -0.1in
    \end{footnotesize}
\end{table}

\subsubsection{RMTPP Configurations}
RMTPP's hyperparameter settings are presented in \cref{tab:rmtpp_hyp}. The three values of "MS" refers to the number of dimensions for time embeddings (used by history encoders), for outputs of the history encoder, and for the history embeddings, respectively. The hyperparameters of RMTPP are more conservative because we notice that a large learning rate or training step always leads to training failure. 

%The original RMTPP is a TPP model. To enable true MTPP support, we duplicated the predefined intensity function, one for each marker, and train this patched RMTPP on all real-world datasets using \cref{eqn:mtpp_loss}.

\begin{table}[!ht]
    \centering
    \caption{Hyperparameter settings for RMTPP.}
    \label{tab:rmtpp_hyp}
    \vskip 0.15in
    \begin{footnotesize}
    \begin{tabular}{ccccc}
    \toprule
      Datasets & Steps & MS & BS & LR \\
    \midrule
      Retweet & [20,000, 100,000] & [32, 32, 32] & 128 & 0.002 \\
      MOOC & [80,000, 400,000] & [48, 32, 32] & 32 & 0.002 \\
      SO & [2,000, 10,000] & [32, 32, 16] & 32 & 0.001 \\
      BO & [1,000, 2,500] & [48, 32, 32] & 8 & 0.001 \\
      Synthetic & [1,000, 10000] & [32, 32, 16] & 128 & 0.002 \\
    \bottomrule
    \end{tabular}
    \vskip -0.1in
    \end{footnotesize}
\end{table}

\subsubsection{LogNormMix Configurations}
%\cref{tab:lognormmix_hyp} provides the hyperparameters used when training LogNormMix on synthetic and real-world datasets. The three values of "MS" are the number of dimensions for the history embedding and mark embeddings, and the number of mixed log-norm distributions.
\cref{tab:lognormmix_hyp} provides the hyperparameters in LogNormMix. The three values of "MS" are the number of dimensions for the history embedding and mark embeddings, and the number of mixed log-norm distributions.

\begin{table}[!ht]
    \centering
    \caption{Hyperparameter settings for LogNormMix.}
    \label{tab:lognormmix_hyp}
    \vskip 0.15in
    \begin{footnotesize}
    \begin{tabular}{ccccc}
      \toprule
      Datasets & Steps & MS & BS & LR \\
    \midrule
      Retweet & [80,000, 400,000] & [32, 32, 16] & 32 & 0.002 \\
      MOOC & [80,000, 400,000] & [32, 32, 32] & 32 & 0.002 \\
      SO & [40,000, 200,000] & [32, 48, 32] & 32 & 0.002 \\
      BO & [4,000, 20,000] & [32, 32, 32] & 8 & 0.002 \\
      Synthetic & [1,000, 10,000] & [32, 32, 64] & 128 & 0.002 \\
      \bottomrule
    \end{tabular}
    \end{footnotesize}
    \vskip -0.1in
\end{table}

\subsubsection{THP Configurations}
\cref{tab:thp_hyp} shows all hyperparameter settings of THP. % when trained on synthetic and real-world datasets. 
The six values of "MS" are the number of dimensions for the Transformer's input vectors, the hidden outputs from an RNN which is on top of the Transformer encoder, the vectors used by self-attentions(\(q\), \(k\), and \(v\)), the number of Transformer layers, and heads.

\begin{table}[!ht]
    \centering
    \caption{Hyperparameter settings for THP.}
    \label{tab:thp_hyp}
    \vskip 0.15in
    \begin{footnotesize}
    \begin{tabular}{ccccc}
    \toprule
      Datasets & Steps & MS & BS & LR \\
    \midrule
      Retweet & [40,000, 200,000] & \makecell{[16, 16, 32, 8, 3, 3]} & 32 & 0.002 \\
      MOOC & [80,000, 400,000] & \makecell{[16, 16, 32, 16, 3, 3]} & 32 & 0.002 \\
      SO & [80,000, 400,000] & \makecell{[16, 16, 32, 8, 3, 3]} & 32 & 0.002 \\
      BO & [4,000, 20,000] & \makecell{[16, 16, 32, 8, 3, 4]} & 8 & 0.002 \\
      Synthetic & [1,000, 10,000] & \makecell{[16, 32, 64, 16, 3, 4]} & 128 & 0.002 \\
    \bottomrule
    \end{tabular}
    \vskip -0.1in
    \end{footnotesize}
\end{table}

\subsubsection{SAHP Configurations}
The hyperparameter settings for SAHP are available in \cref{tab:sahp_hyp}. The first six values of "MS" share the same meaning as in THP, while the last is the dropout rate.

\begin{table}[h]
    \centering
    \caption{Hyperparameter settings for SAHP.}
    \label{tab:sahp_hyp}
    \vskip 0.15in
    \begin{footnotesize}
    \begin{tabular}{ccccc}
    \toprule
      Datasets & Steps & MS & BS & LR \\
    \midrule
      Retweet & [40,000, 200,000] & \makecell{[16, 16, 32, 8, 3, 3, 0.1]} & 32 & 0.002 \\
      MOOC & [80,000, 400,000] & \makecell{[16, 16, 32, 16, 3, 3, 0.1]} & 32 & 0.002 \\
      SO & [80,000, 400,000] & \makecell{[16, 16, 32, 8, 3, 3, 0.1]} & 32 & 0.002 \\
      BO & [4,000, 20,000] & \makecell{[16, 16, 32, 8, 3, 4, 0.1]} & 8 & 0.002 \\
      Synthetic & [1,000, 10,000] & \makecell{[16, 32, 64, 16, 3, 4, 0.1]} & 128 & 0.002 \\
    \bottomrule
    \end{tabular}
    \vskip -0.1in
    \end{footnotesize}
\end{table}

\subsubsection{IFIB-N Configurations}
The hyperparameter settings for IFIB-N are available in \cref{tab:ifib_n_hyp}. The four numbers of "MS" mean the dimension of the historic space, the dimension of the conditional probability, namely \(\mathbf{u}_i\) and \(\mathbf{v}_i\) in \cref{fig:cifib}, the dimension of \(\mathbf{f}(x)\), and the number of non-negative layers in IEM (Integral Estimation Module). We find that probability distribution \(p^*(d_i|t, d_1, \cdots, d_{i-1})\) might be drastically different from \(p^*(t)\), so we allocate two IEMs: one for calculating \(p^*(t)\), and the other for \(p^*(d_i|t, d_1, \cdots, d_{i-1})\). %The scaling factor \(\alpha\) and \(\beta\) are fixed to \(0.5\) and \(0.1\) in all IFIB-N models.
\begin{table}[ht]
    \centering
    \caption{Hyperparameter settings for IFIB-N.}
    \label{tab:ifib_n_hyp}
    \vskip 0.15in
    \begin{footnotesize}
    \begin{tabular}{ccccc}
    \toprule
      Datasets & Steps & MS & BS & LR \\
    \midrule
      COVID-19 & [20,000, 100,000] & \makecell{[32, 32, 16, 3]} & 32 & 0.002 \\
      Citibike & [20,000, 100,000] & \makecell{[32, 32, 16, 3]} & 32 & 0.002 \\
      Earthquakes & [20,000, 100,000] & \makecell{[32, 32, 16, 3]} & 32 & 0.002 \\
      Synthetic & [1,000, 10,000] & \makecell{[32, 32, 64, 3]} & 128 & 0.002 \\
    \bottomrule
    \end{tabular}
    \vskip -0.1in
    \end{footnotesize}
\end{table}

% I will move this part to the appendix. %This fact makes results in \cref{sec:mtpp_et_performance} precious. However, %, also ubiquitous in the real world, %One might notice that some results in \cref{tab:IFIB_real_world_mae_e} look suspicious: why do we report ridiculously large MAE-E values for FullyNN and FENN on MOOC and Bookorder? why does SAHP perform relatively well on StackOverflow and Bookorder dataset while its MAE-E@\(50\%\)(Q2) and MAE-E@\(75\%\)(Q3) on MOOC is bigger than 500,000(In fact, that two value is 826129.25 and 968786.0)?
\section{Analysis on Event-time Prediction with IFIB-C}
\label{app:mae_e_analysis}
The time-event task has been well-studied while the event-time task is barely discussed. To have a better understanding of the results in \cref{tab:IFIB_real_world_mae_e}, we review how event mark and time are predicted in the event-time prediction task. First, we recall \cref{eqn:pred_time}:
\begin{equation}
\begin{aligned}
%    m_p &= \argmax_{m}{p^*(m)} = \frac{\operatorname{IFIB}(m, t_l, \mathcal{H}_{t_l})}{\sum_{n \in M}{\operatorname{IFIB}(n, t_l, \mathcal{H}_{t_l})}} \\
\label{eqn:pred_time_in_app}
    P^*(t|m) = \frac{P^*(m, t)}{P^*(m)} = \frac{1}{\int_{t_l}^{+\infty}{p^*(m, \tau)d\tau}}\int_{t_l}^{t}{p^*(m, \tau)d\tau} \\
\int_{t_l}^{t}{p^*(m, \tau)d\tau} = \int_{t_l}^{+\infty}{p^*(m, \tau)d\tau} - \int_{t}^{+\infty}{p^*(m, \tau)d\tau}
\end{aligned}
\end{equation}
%where \(p^*(m, t)\) is the probability distribution of a marked TPP. 
As indicated in \cref{sec:ifib-based_solution}, we identify mark \(m_p\) which leads to the largest \(\int_{t_l}^{+\infty}{p^*(m, \tau)d\tau}\), and the corresponding time \(t_p\) which is the minimum time satisfying \(P^*(t|m_p) \geq 0.5\). So, accurate estimation of \(\int_{t_l}^{+\infty}{p^*(m, \tau)d\tau}\) is essential. As shown in \cref{eqn:pred_time_in_app}, \(P^*(m) = \int_{t_l}^{+\infty}{p^*(m, \tau)d\tau}\) and it must satisfy:
\begin{equation}
\sum_{\mathrm{M}}{P^*(m) = 1}
\end{equation}
As stated in \cref{sec:method}, IFIB framework can easily enforce $\sum_{\mathrm{M}}{P^*(m) = 1}$. In contrast, the marked TPP models (FullyNN, FENN, SAHP, THP) which are capable of the event-time prediction task fail to do so. \cref{tab:IFIB_real_world_sum_of_pm} shows $\sum_{\mathrm{M}}{P^*(m)}$ estimated by different models. Interestingly, we notice that a close-to-one sum always indicates a better MAE-E in \cref{tab:IFIB_real_world_mae_e}. For example, IFIB and SAHP manage to keep the sum around 1 on the Bookorder dataset, while FENN and FullyNN do not. In \cref{tab:IFIB_real_world_mae_e}, IFIB and SAHP's MAE@\(25\%\), MAE@\(50\%\) ,and MAE@\(75\%\) on bookorder are marginally better than FullyNN and FENN. This help explain why IFIB-C has an outstanding performance in \cref{tab:IFIB_real_world_mae_e}.

\begin{table}[!ht]
    \caption{$\sum_{n \in \mathrm{M}}{P^*(m)}$ by IFIB, FullyNN, FENN, SAHP, and THP.}
    \centering
    \begin{small}
\vskip 0.15in
    \begin{tabular}{lcccc}
        \toprule
                                     & Retweet & SO & MOOC & BO \\
        \toprule
        IFIB(Ours)                   & \textbf{1.0000\tiny{\(\pm\)0.0000}} & \textbf{1.0000\tiny{\(\pm\)0.0000}} &\textbf{1.0000\tiny{\(\pm\)0.0000}} & \textbf{1.0000\tiny{\(\pm\)0.0000}} \\
        FullyNN                      & 1.2760\tiny{\(\pm\)0.2266} & 1.0730\tiny{\(\pm\)0.0048} & 8.2638\tiny{\(\pm\)0.1903} & 0.6907\tiny{\(\pm\)0.0054} \\
        FENN                         & 1.1862\tiny{\(\pm\)0.1354} & 0.9977\tiny{\(\pm\)0.0160} & 5.0623\tiny{\(\pm\)0.0582} & 0.6952\tiny{\(\pm\)0.0081} \\
        SAHP                         & 3.3466\tiny{\(\pm\)0.2098} & 1.0059\tiny{\(\pm\)0.0001} & 3108.4\tiny{\(\pm\)2617.9} & 1.0526\tiny{\(\pm\)0.0018} \\
        THP                          & / & 1.0043\tiny{\(\pm\)0.0000} & / & / \\
        \bottomrule
    \end{tabular}
    \end{small}
    \label{tab:IFIB_real_world_sum_of_pm}
\vskip -0.1in
\end{table}

% \subsection{The numerical Integration}
Now, we explain why FullyNN, FENN, SAHP, and THP cannot enforce $\sum_{\mathrm{M}}{P^*(m) = 1}$. FullyNN, FENN, SAHP, and THP use numerical integration methods, such as the Monte-Carlo integration, to calculate the cumulative distribution function \(P^*(m, t)\) because these models cannot provide closed-form probability distribution \(p^*(m, t)\). However, numerical integration methods focus on calculating the definite integral. This means that the integration interval must have a finite length. However, \(P^*(m) = \int_{t_l}^{+\infty}{p^*(m, \tau)d\tau}\) so that the numerical integration methods are not suitable.

What about replacing the positive infinity in \(P^*(m)\) with a relative huge number \(L\) and calculating the integral of \(p^*(m, t)\) over \([t_l, t_l + L]\)? This idea is feasible because the computer never truly knows infinity. A standard solution is to set up a large number to imitate infinity\footnote{In fact, we replace the positive infinity in \(P^*(m)\) with \(\min(\bar{\Delta t} + 10\sigma_{\Delta t}, 10^6)\) in our implementation. \(\bar{\Delta t}\) and \(\sigma_{\Delta t}\) are the average and standard deviation of time intervals, respectively. \(\hat{\Delta t} + 10\sigma_{\Delta t}\) is large enough that no existing observation is larger than it.}. %However, we encounter another issue: how many samples do we need in \([t_l, t_l + L]\)?

\begin{figure*}[ht]
    \centering
    \begin{subfigure}{0.45\textwidth}
        \includegraphics[width=\textwidth]{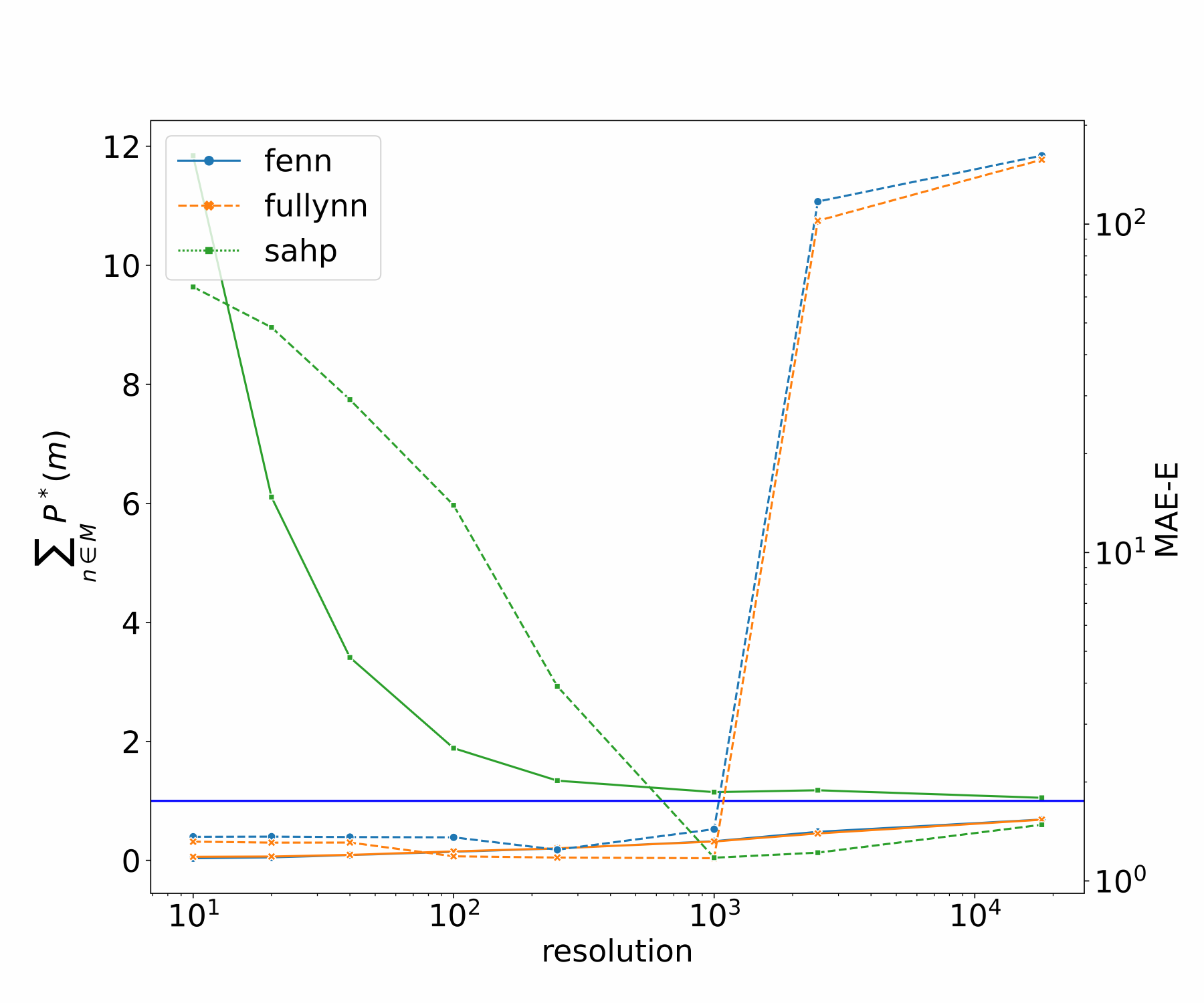}
        \label{fig:res_bookorder}
        \caption{Bookorder}
    \end{subfigure}
    \begin{subfigure}{0.45\textwidth}
        \includegraphics[width=\textwidth]{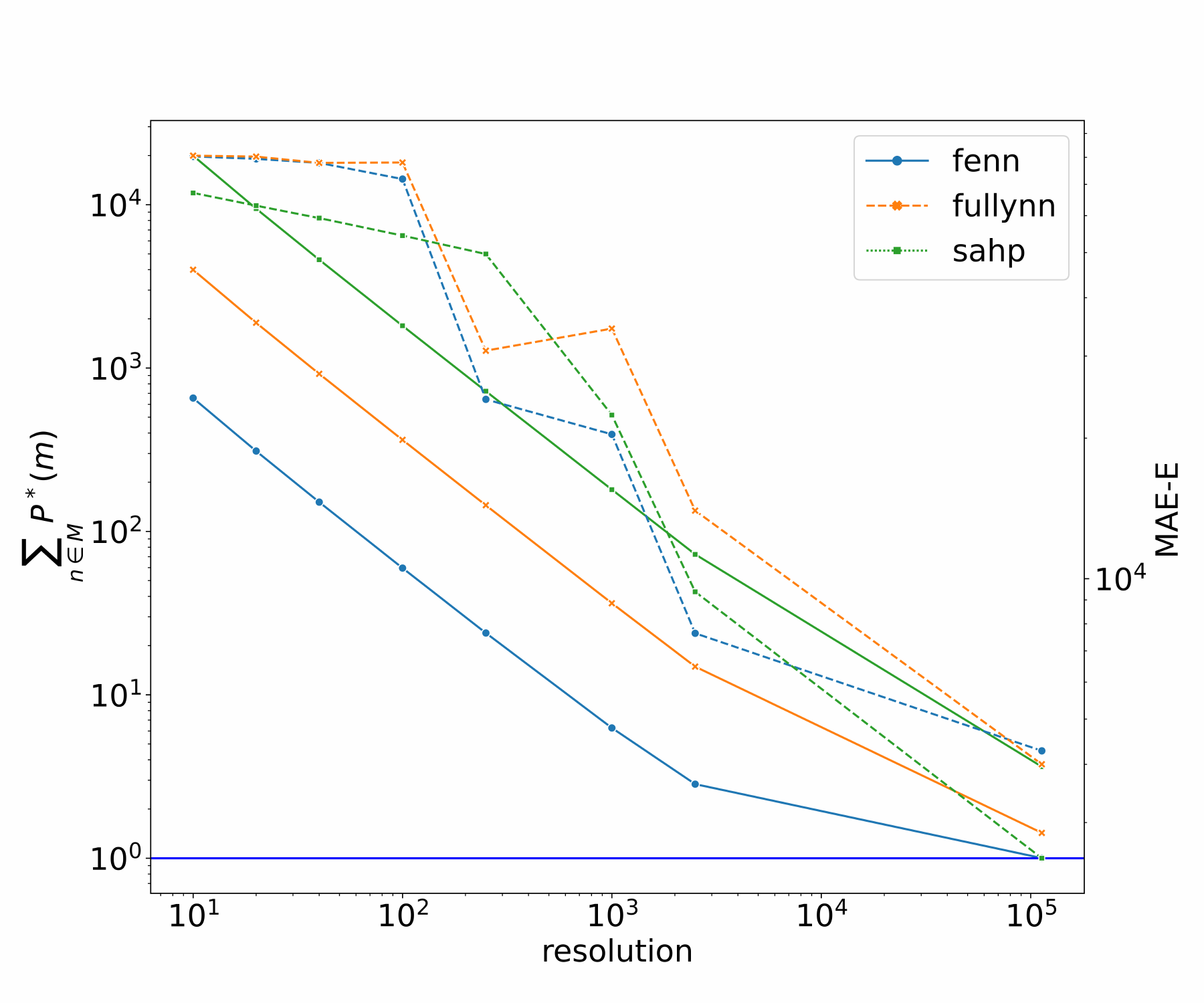}
        \label{fig:res_retweet}
        \caption{Retweet}
    \end{subfigure}
    \begin{subfigure}{0.45\textwidth}
        \includegraphics[width=\textwidth]{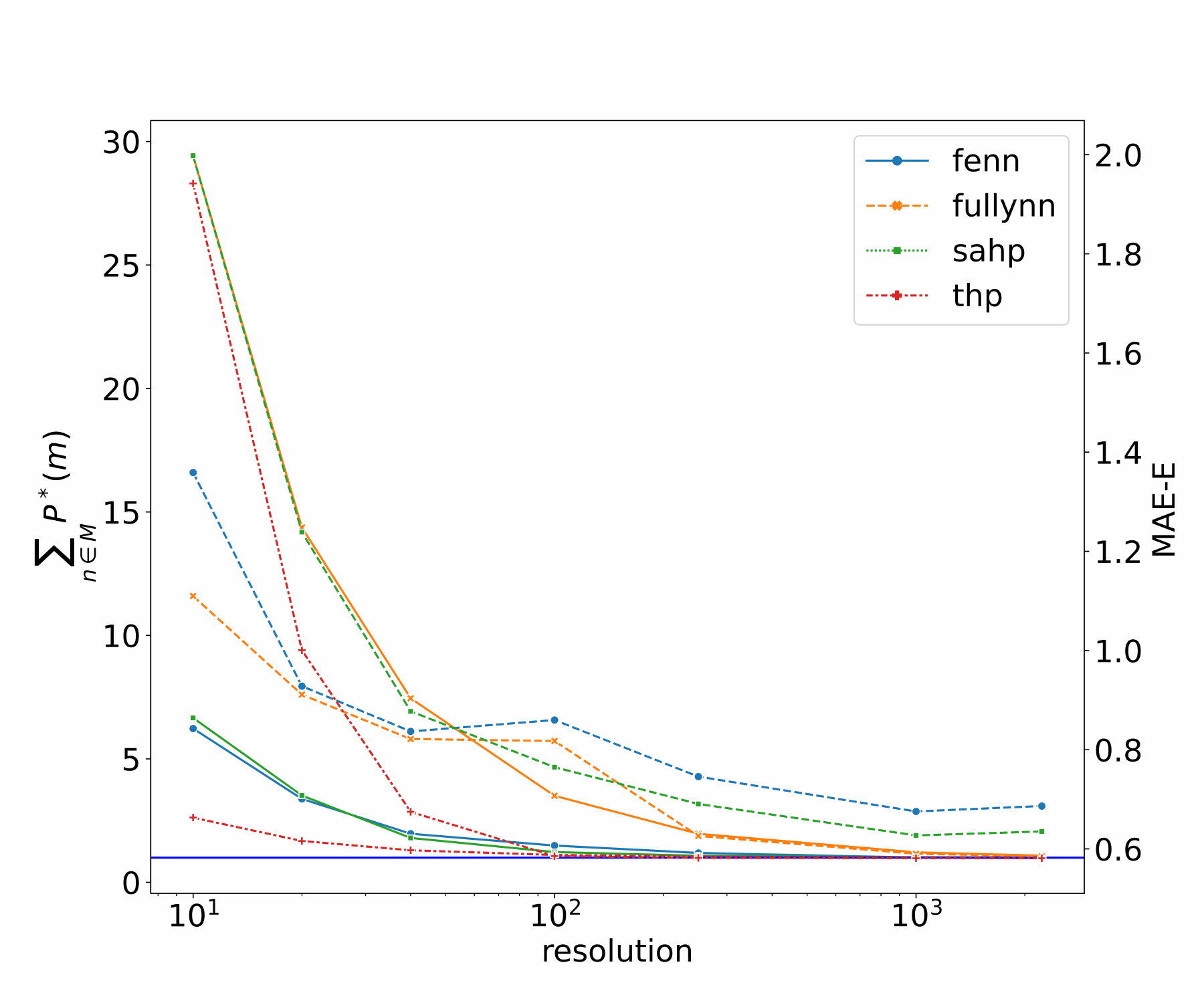}
        \label{fig:res_stackoverflow}
        \caption{StackOverflow}
    \end{subfigure} 
    \caption{The value of \(\sum_{\mathrm{M}}{P^*(m)}\) with different numbers of sampling points. The graph on MOOC is unavailable because of memory issues. The blue line is \(y = 1\). Solid lines refer to the sum of \(P^*(m)\), and dashed lines refer to the MAE-E value with the corresponding sampling rate. These three graphs prove that we need plenty of sampling points for calculating every \(P^*(m)\), and more accurate \(P^*(m)\) could often result in sensible time predictions for event-time prediction tasks.}
    \label{fig:probe_res}
\end{figure*}

\begin{table}[!ht]
    \caption{\(L\) and the number of samples of each real-world dataset used in experiments}
    \centering
    \begin{small}
\vskip 0.15in
    \begin{tabular}{lcccc}
        \toprule
                                     & Retweet & SO & MOOC & BO \\
        \toprule
        Average of \(\Delta t\)      & 2550.2   & 0.8167 &   16865   &   1.3273   \\
        Standard Deviation of \(\Delta t\)&16230& 1.0333 &   95194   &   20.240   \\
        \(L\)                        & 164847   & 11.150 &   968807  &   203.73   \\
        Number of samples(Average)   & 113048   & 2229   &   16819   &   18094    \\
        Number of samples(Maximum)   & 200000   & 2229   &   77319   &   18094    \\
        Number of samples(Minimum)   & 37878    & 2229   &    781    &   18094    \\
        Average gap between samples  & 1.4582   & 0.005  &  57.602   &   0.0113   \\
        \bottomrule
    \end{tabular}
    \end{small}
    \label{tab:l_and_sample}
\vskip -0.1in
\end{table}
\cref{tab:l_and_sample} shows the value of \(L\) and the number of samples we use during evaluation\footnote{We limit the number of samples on Retweet and MOOC because of the memory limitation. Specifically, we force the production of sequence length, the number of samples, and the number of marks must not exceed 30,000,000. Batch size is not included because it is always 1 during evaluation.}. According to \cref{tab:IFIB_real_world_sum_of_pm}, $\sum_{\mathrm{M}}{P^*(m)}$ on Stackoverflow is much closer to 1 compared with other datasets. The reason is that the sampling points on Stackoverflow are much denser, so the numerical estimation of \(P^*(m)\) is accurate. On the other hand, insufficient sampling points harm the accuracy of \(P^*(m)\) estimation. For example, the average gap between samples on MOOC is 57.602, over 10,000 times larger than Stackoverflow's. Such a colossal gap causes $\sum_{\mathrm{M}}{P^*(m)}$ on SAHP to be 1048.2, leading to ridiculous time predictions. Further experiment results about the relation between the number of samples and the sum of \(P^*(t)\) are reported in \cref{fig:probe_res}. Forcing the sampling gap to 0.005 on Retweet and MOOC is also impossible. Basic calculation tells us that we will have 32,969,400 for Retweet or 193,761,400 sampling points for MOOC to calculate one \(P^*(m)\). No modern computation device can handle that huge amount of calculation.\par

One could notice that in \cref{fig:probe_res}, FullyNN and FENN give poor time predictions on Bookorder when the number of sampling points is bigger and \(\sum_{\mathrm{M}}{P^*(m)}\) is closer to 1. After investigating the learning distribution over time, we find these two models might overfit the training set. Such overfitting introduces super high (over 200) yet narrow (around 0.01) spikes into the intensity function, which can only catch with a huge sampling rate. Such spikes severely damage the time prediction performance of FENN and FullyNN on Bookorder. This problem further proves the superiority of the IFIB framework.

% \subsection{FENN and FullyNN}

% We discover that as discussed in \cref{sec:IFIBP}, IFIB possesses the property naturally. SAHP and THP hold the property in theory, but the numerical method (like the Monte Carlo) might be inaccurate. In practice, this is not always possible, as shown in \cref{tab:IFIB_real_world_sum_of_pm}. Second, FullyNN and FENN have the intrinsic weakness to satisfy the property as discussed in \cref{apx:ETE}. This is consistent with the results in \cref{tab:IFIB_real_world_sum_of_pm}. Moreover, after comparing FENN and FullyNN's performance on Stackoverflow with Retweet and Bookorder, we conclude that such weakness is responsible for their poor MAE-E in \cref{tab:IFIB_real_world_sum_of_pm}. 

% Do we need this part?
% Yes, we need them.
\section{IFIB-C Result Visualization}
\begin{figure*}[ht]
    \centering
    \begin{subfigure}{1.05\textwidth}
        \includegraphics[width=\textwidth]{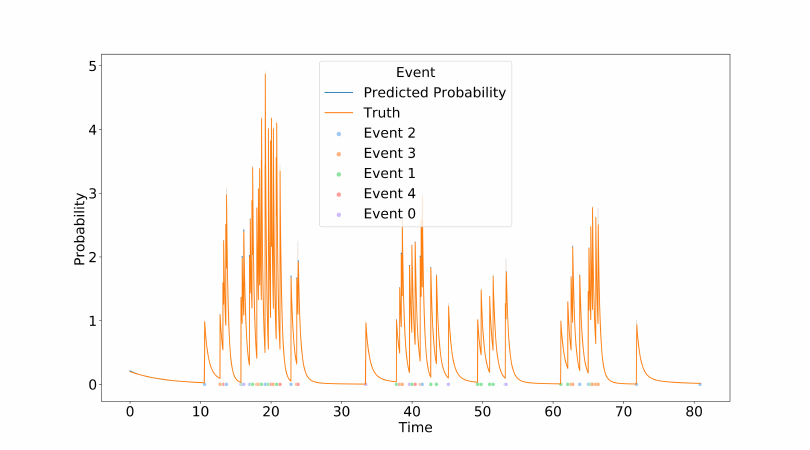}
        \label{fig:h1}
        \caption{Synthetic dataset - Hawkes\_1 (Spearman coefficient \(\rho = 1.0000\), \(L_1 = 0.1301\))}
    \end{subfigure}
    \begin{subfigure}{1.05\textwidth}
        \includegraphics[width=\textwidth]{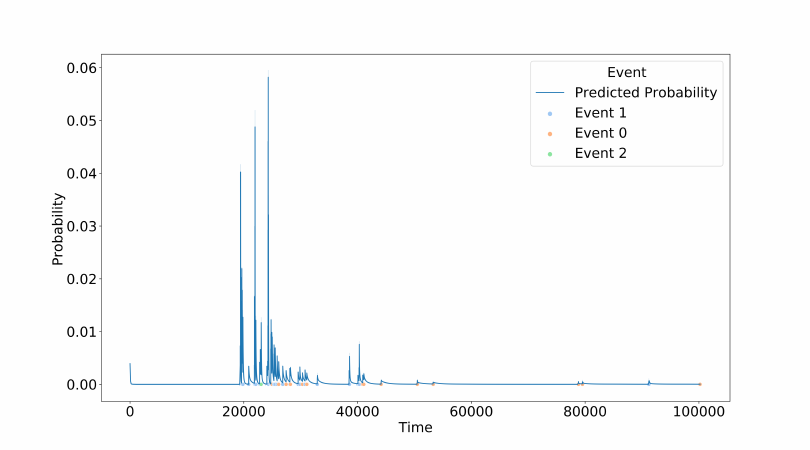}
        \label{fig:h2}
        \caption{Real-world dataset - Retweet}
    \end{subfigure}
    \caption{The comparison between the probability distribution learned by IFIB-C and the ground truth.}
\end{figure*}

%\subsection{Synthetic datasets}
\label{app:plot_synthetic}
We visualize the comparison between predicted \(\hat{p}^*(m, t)\) using IFIB-C and the ground truth probability distributions \(p^*(m, t)\) on synthetic dataset, Hawkes\_1, in \cref{fig:h1}. The Spearman coefficient \(\rho\) and \(L^1\) distance between \(\hat{p}^*(m, t)\) and \(p^*(m, t)\) demonstrate the desirable performance of IFIB-C. We also visualize the predicted \(\hat{p}^*(m, t)\) using IFIB-C in \cref{fig:h2}. We can observe the consistency between \(\hat{p}^*(m, t)\) and the occurrence frequency of events along with time.

\end{document}